\documentclass{article}

\PassOptionsToPackage{numbers, compress}{natbib}
\usepackage[preprint]{neurips_2026}

\usepackage[utf8]{inputenc} 
\usepackage[T1]{fontenc}    
\usepackage{hyperref}       
\usepackage{url}            
\usepackage{booktabs}       
\usepackage{amsfonts}       
\usepackage{nicefrac}       
\usepackage{microtype}      
\usepackage{xcolor}         
\usepackage{subcaption}

\usepackage{float}
\usepackage{amsmath}
\usepackage{amssymb}   
\usepackage{mathtools}
\usepackage[capitalise]{cleveref}
\usepackage{bm}      
\usepackage{tabularx}
\usepackage{caption}
\usepackage{wrapfig}
\usepackage{adjustbox}
\usepackage{enumitem}
\usepackage[table]{xcolor}
\usepackage{xspace}
\usepackage{etoc}

\newcommand{\algname}{FlowBender\xspace}

\usepackage{algorithm}
\usepackage[noend]{algpseudocode}
\definecolor{commentcolor}{RGB}{70, 110, 140}

\definecolor{algcomment}{RGB}{120, 120, 120}

\title{\algname: Feedback-Aware Training for Self-Correcting Conditional Flows}

\author{%
  Daniel Gilo\textsuperscript{1} \quad 
  Sven Elflein\textsuperscript{2,3,4} \quad 
  Ido Sobol\textsuperscript{1} \quad 
  Or Litany\textsuperscript{1,2} \\
  \textsuperscript{1}Technion \quad 
  \textsuperscript{2}NVIDIA \quad 
  \textsuperscript{3}University of Toronto \quad
  \textsuperscript{4}Vector Institute \\
  \texttt{danielgilo@cs.technion.ac.il}
}

\begin{document}

\maketitle

\begin{abstract}
    Conditional diffusion and flow models routinely fail to satisfy the very constraints that define their task. For instance, a depth-conditioned model often produces images whose re-extracted depth disagrees with the input, even though the forward operator---the depth predictor defining the constraint---is available during both training and inference. Existing approaches generally fall into two categories: supervised models that treat the conditioning signal as a static cue and ignore alignment information at inference, and guidance-based methods that consult it through hand-tuned linear updates, typically trading fidelity to the condition against the plausibility of the generated sample. We argue that the fundamental gap in both paradigms is that the model is never trained to utilize its own alignment error. We introduce \algname, a closed-loop framework that treats this error as a first-class input, training the network to learn a correction policy conditioned on inference-time feedback. At each step, an unguided look-ahead pass estimates the clean signal, a task-specific deviation is computed via the forward operator, and a refinement pass consumes this signal to produce a corrected velocity. We propose several variants of \algname, including a gradient-based formulation for differentiable operators and a zero-order variant for non-differentiable settings such as JPEG compression. For efficient sampling, we introduce a prior-step shortcut that enables closed-loop correction at a minimal additional computational cost. Across image-to-image translation, restoration, and 3D mesh texturing, \algname consistently outperforms standard supervised baselines, alignment-loss-augmented training, and state-of-the-art inference-time guidance, improving fidelity and plausibility simultaneously rather than trading them against each other. Project page: \url{https://flow-bender.github.io/}.
\end{abstract}

\section{Introduction}
\label{sec:introduction}

Diffusion and Flow Matching (FM) models \cite{sohl2015deep, ho2020denoising, lipman2022flow} have become the dominant paradigm for generative modeling, and a primary use-case is \emph{conditional} generation: producing samples $\mathbf{x}$ aligned with an external signal $\mathbf{y}$, $e.g.,$ a text prompt, a depth map, or a geometric constraint. Effective conditional sampling requires both \emph{fidelity} to $\mathbf{y}$ and \emph{plausibility} with respect to the target data manifold. Existing methods, however, routinely fail on the first axis: for instance, a ControlNet \cite{zhang2023adding} conditioned on depth or edge maps often produces images whose re-extracted measurements disagree with the input (Fig.~\ref{fig:image_to_image}). This inconsistency occurs even though the forward operator $\mathcal{H}$ that relates the sample to the conditioning signal, such as a depth predictor, edge detector, or renderer, is typically available during both training and inference.

This failure mode reveals a missed opportunity. Supervised conditional models \cite{nichol2021glide, saharia2022palette, rombach2022high, zhang2023adding} treat $\mathbf{y}$ as a static cue and operate as \emph{open-loop} systems at inference: even when the evolving sample drifts from the constraint, and even when the operator $\mathcal{H}$ is available to compute a deviation signal, the network has no built-in mechanism to consult this feedback and adjust its trajectory. The very alignment information that motivates the task is left on the table.

\begin{figure}
    \centering
    \includegraphics[width=1\linewidth]{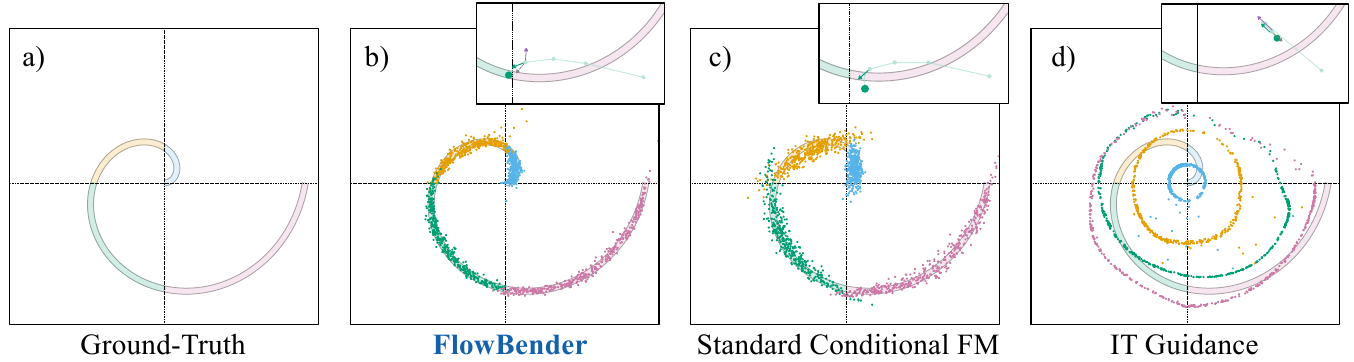}
\caption{A conditional flow model is trained to sample from the 2D Archimedean spiral distribution shown in (a), which is partitioned by quadrant into four classes representing distinct radius ranges. (b--d) point colors denote the provided condition target classes. (b) \textbf{\algname} learns to internalize feedback from a radial guidance signal, achieving faithful alignment with both class constraints and the data manifold. (c) \textbf{Standard conditional generation} often violates class boundaries and misses the target distribution. (d) \textbf{Inference-time (IT) Guidance} satisfies radial constraints but drives samples off-manifold entirely. \textbf{Insets} showcase representative sampling trajectories for a green-class target, focusing on a specific velocity prediction at $t=0.7$. The grey arrows (b, d) represent the first-pass or unconditional prediction, the purple arrow is the guidance component, and the green arrow is the final step taken. In (d), guidance dominates, pushing the final sample (large green dot) off-manifold. While the standard model (c) misses both the class and data distribution, our framework (b) learned to aggregate the components to arrive safely at the correct class within the data manifold. See Appendix \ref{appendix:2d_toy} for further details.
}
    \label{fig:toy_exp}
\end{figure}

Guidance-based methods \cite{dhariwal2021diffusion, chung2022diffusion, bansal2023universal, patel2025flowchef} do consult $\mathcal{H}$ at inference, steering the trajectory using the gradient
of a measurement-matching objective. 
However, these approaches utilize feedback only during inference, through a manually-tuned linear update rule which creates a fundamental train-test discrepancy and a delicate tuning problem: too little guidance fails to satisfy the constraint, too much pushes the trajectory off the data manifold \cite{he2023manifold, ye2024tfg}. 
Figure~\ref{fig:toy_exp} illustrates both failure modes on a 2D Archimedean spiral: standard conditional generation drifts across class boundaries (c), while inference-time guidance enforces the radial constraint at the cost of missing the data manifold (d).

We argue the fundamental gap in both paradigms is that the model is never \emph{trained} to utilize its own alignment error. We close this gap by treating this error as a first-class input and training the model to learn a correction policy over it.
Concretely, at each sampling step, the model first performs an unguided look-ahead pass to estimate the clean signal, from which we derive a task-specific deviation signal via the operator $\mathcal{H}$. 
A second refinement pass then consumes this error 
alongside the standard inputs and emits a corrected update — \emph{bending} the trajectory toward the conditional manifold (Fig.~\ref{fig:toy_exp}(b), Fig.~\ref{fig:method}).  We refer to our framework as \textbf{\algname} -- a closed-loop system that self-corrects throughout sampling. 

The framework has three notable properties. First, the learned correction is \emph{not} guidance with better hyperparameters: orthogonal decomposition shows that 80\% of the second-pass correction energy lies orthogonal to the gradient, indicating that the model exploits feedback through a non-linear policy that scalar-weighted schemes cannot express (\S\ref{sec:analysis}). Second, \algname does not require the error signal to be a gradient: a \emph{zero-order variant} feeds the raw measurement-space error directly to the network, extending correction to non-differentiable or black-box operators, such as JPEG compression and third-party APIs, where gradient guidance is inapplicable. Third, a \emph{prior-step shortcut} allows reducing inference cost to as little as $N{+}1$ evaluations for $N$-step sampling, nearly matching open-loop efficiency while retaining corrective benefits (\S\ref{sec:inference}).

Empirically, \algname consistently outperforms standard supervised training, alignment-loss-augmented training \cite{li2024controlnet++}, and state-of-the-art inference-time guidance \cite{patel2025flowchef} across 3D texturing, JPEG restoration, and image translation (super-resolution, depth/edge-to-RGB). Remarkably, while our closed-loop framework is designed to enhance fidelity, it consistently improves plausibility as well (e.g., FID), in direct contrast to the fidelity–plausibility trade-off that manifests in traditional guidance approaches.

Our main contributions are:

\begin{itemize}[nosep, leftmargin=*]
    \item We present \algname, a closed-loop approach for conditional diffusion and FM models that replaces hand-tuned guidance updates with a learned policy over the model's own alignment error. The framework is architecturally agnostic and integrates with existing adapters such as ControlNet \cite{zhang2023adding} and LoRA \cite{hu2022lora}.

    \item We include a zero-order variant that extends learned correction to non-differentiable or black-box operators, a regime where gradient-based guidance is inapplicable.
    \item We propose a prior-step shortcut for closed-loop correction at near-open-loop inference cost.  
    \item We demonstrate simultaneous gains in fidelity and plausibility across image translation, restoration, and 3D texturing; analysis attributes these results to a learned correction policy that utilizes feedback non-linearly, in contrast to the rigid scalar-weighting of traditional guidance.
\end{itemize}

\section{Related Work}
\label{sec:related_work}

\paragraph{Conditional Sampling via Open-Loop Training.}
Conditional diffusion and flow-matching models typically parameterize a score function or velocity field given a static conditioning signal. For high-dimensional domains like images or meshes, foundational models \cite{rombach2022high, esser2024scaling, flux2024, flux-2-2025} require massive paired data; specific controls (e.g., depth or masks) are thus often added via adapters like ControlNet \cite{zhang2023adding} or LoRA \cite{hu2022lora}. Although theoretically sampling from the posterior $p(\mathbf{x} \mid \mathbf{y})$ \cite{batzolis2021conditional}, these models often fail to satisfy conditioning constraints in practice \cite{ho2022classifier, karras2024guiding, sobol2024zero}. We identify a limitation in treating $\mathbf{y}$ as a static hint: the model lacks a mechanism to evaluate or adjust its trajectory even though alignment diagnostics are often available during both training and inference. 
Self-Conditioning~\cite{chen2022analog} is a related technique that feeds the model's prior-step sample prediction back into the network as an additional condition to improve overall sample quality. However, like other recent enhanced training schemes \cite{li2024controlnet++, xu2024ctrlora, luo2024readout, elata2025invfusion}, this approach remains fundamentally open-loop: it does not provide the model with an estimate of its deviation with respect to $\mathbf{y}$. By introducing alignment error as an explicit input, we transform the generation process into a closed-loop system capable of active self-correction.

\vspace{-4pt}
\paragraph{Conditional Sampling via Bayesian Guidance.} 
An alternative paradigm treats conditional sampling as Bayesian posterior inference, decomposing the conditional score into prior and likelihood terms:
\begin{equation}
    \nabla_{\mathbf{x}_t} \log p_t(\mathbf{x}_t \mid \mathbf{y}) = \nabla_{\mathbf{x}_t} \log p_t(\mathbf{x}_t) + \nabla_{\mathbf{x}_t} \log p_t(\mathbf{y} \mid \mathbf{x}_t).
    \label{eq:bayes_posterior}
\end{equation}
While the prior is typically approximated via a pre-trained denoiser, the likelihood term is defined by the intractable integral $p_t(\mathbf{y} \mid \mathbf{x}_t) = \int p(\mathbf{y} \mid \mathbf{x}_1) p(\mathbf{x}_1 \mid \mathbf{x}_t) \mathrm{d}\mathbf{x}_1$. Paradigms for approximating the likelihood vary: Classifier Guidance \cite{dhariwal2021diffusion} utilizes time-dependent classifiers, Classifier-Free Guidance (CFG) \cite{ho2022classifier} leverages the difference between conditional and unconditional score estimates, and training-free methods \cite{daras2024survey, kawar2021snips, kawar2022denoising, chung2022diffusion,chung2022improving,  song2023pseudoinverse, wang2022zero, zhang2025improving, bansal2023universal, patel2025flowchef, kim2025flowdps} approximate it using distance metrics between the predicted clean signal $\hat{\mathbf{x}}_1(\mathbf{x}_t)$ and the measurement $\mathbf{y}$. These modular approaches rely on \emph{heuristic} weighting, creating a trade-off between constraint satisfaction and sampling artifacts or divergence \cite{he2023manifold, ye2024tfg}. This standard decomposition involves cascading approximations (estimated prior, point-estimate likelihood, and manual weighting), yielding a suboptimal conditional score. Unlike works that learn CFG coefficients \cite{galashov2025learn, yehezkel2025navigating}, we propose a general paradigm where the model utilizes alignment error as a first-class input. This enables the network to learn a complex, non-linear policy that compensates for systematic inaccuracies and more faithfully samples from the posterior.

\vspace{-10pt}
\paragraph{Learned Iterative Refinement.}
The paradigm of training neural networks to utilize error signals traces back to \emph{learned optimizers}, where a network is trained to predict weight update rules based on gradient information from a base model \cite{andrychowicz2016learning, wichrowska2017learned, metz2020tasks, harrison2022closer, metz2022velo}. This approach was adapted for inverse problems to iteratively refine estimates by incorporating measurement-space reconstruction error as input \cite{carreira2016human, adler2017solving, adler2018learned, li2018deepim, ma2020deep}. In computer vision, learned iterative refinement has advanced view synthesis \cite{flynn2019deepview} and 3D scene reconstruction \cite{chen2024g3r, kang2025ilrm, liu2025quicksplat, yildirim2026geofusionlrm, liu2026diff3r, chen2026gifsplat, long2026idesplat, xu2025resplat, wen2025life}. By learning to leverage feedback from their own errors, these methods improve fidelity over feed-forward baselines while exceeding the efficiency of per-scene optimization. To our knowledge, we are the first to adapt this error-feedback paradigm to conditional diffusion and flow models.

\section{Preliminaries: Conditional Flow Matching Models}
\label{sec:preliminaries}

Flow Matching (FM) \cite{lipman2022flow} models a probability path $p_t(\mathbf{x}_t)$ that interpolates between a noise distribution $p_0(\mathbf{x}_0) \sim \mathcal{N}(0, \mathbf{I})$ and a target data distribution $p_1(\mathbf{x}_1)$. While the term \emph{conditional flow matching} (CFM) often refers specifically to the training framework where paths are constructed relative to data points $\mathbf{x}_1$, we use it here to denote flow models that receive an external task-specific conditioning signal $\mathbf{c}$, $e.g.$, a corrupted image or depth map. The forward process $p_t(\mathbf{x}_t \mid \mathbf{x}_1)$ for $t \in [0, 1]$ is defined by the linear interpolation:
\begin{equation}
\mathbf{x}_t = a_t \mathbf{x}_1 + \sigma_t \mathbf{x}_0,
\label{eq:forward_process}
\end{equation}
where $a_t$ and $\sigma_t$ are schedule coefficients. This framework typically involves training a network $\mathbf{v}_\theta(\mathbf{x}_t, t, \mathbf{c})$ to approximate the vector field $\mathbf{u}_t(\mathbf{x}_t \mid \mathbf{x}_1) = \dot{a}_t \mathbf{x}_1 + \dot{\sigma}_t \mathbf{x}_0$ via the objective:
\begin{equation}
\mathcal{L}_{\text{FM}} = \mathbb{E}_{t, (\mathbf{x}_1, \mathbf{c}), \mathbf{x}_0} \left[ \| \mathbf{v}_\theta(\mathbf{x}_t, t, \mathbf{c}) - \mathbf{u}_t(\mathbf{x}_t \mid \mathbf{x}_1) \|^2 \right].
\label{eq:loss}
\end{equation}

Notably, $\mathbf{v}_\theta$ provides a principled mechanism to compute a point estimate of the clean signal, $\hat{\mathbf{x}}_1$, at any step $t$. For instance, in optimal transport FM ($a_t=t, \sigma_t=1-t$), the clean estimate is $\hat{\mathbf{x}}_1 = \mathbf{x}_t + (1-t)\mathbf{v}_\theta$.

\section{Feedback-Aware Conditional Flows}
\label{sec:method}

\begin{figure}
    \centering
    \includegraphics[width=1\linewidth]{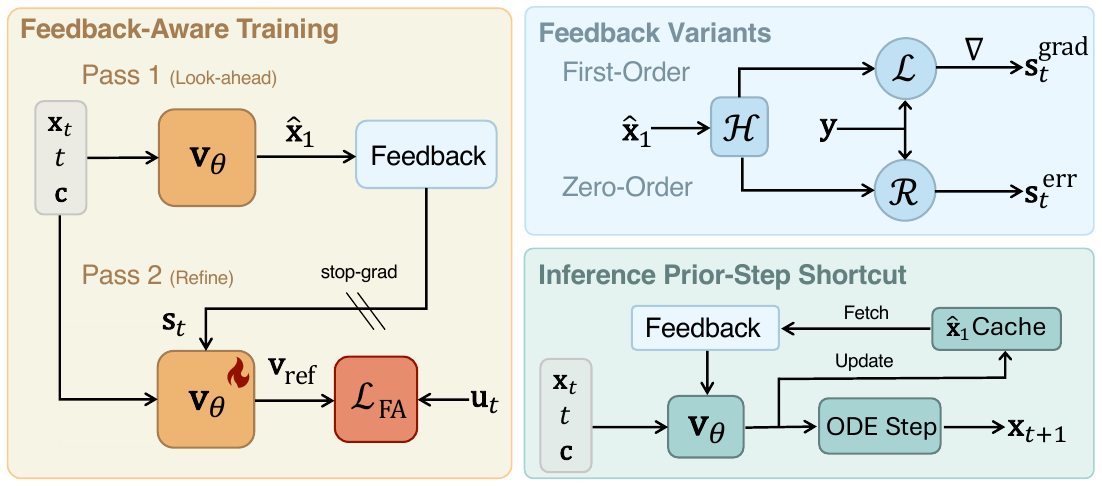}
    \caption{\textbf{\algname overview.} \textbf{(Left)} Training follows a two-pass strategy: a \textit{look-ahead} pass produces a clean-signal estimate $\hat{\mathbf{x}}_1$ to compute the feedback signal $\mathbf{s}_t$, which then conditions a second \textit{refinement} pass. \textbf{(Top-right)} Feedback variants include first-order gradients for differentiable operators and zero-order residuals for non-differentiable or black-box settings. \textbf{(Bottom-right)} At inference, an optional shortcut bypasses the look-ahead pass by fetching $\hat{\mathbf{x}}_1$ from a cached estimate of the previous step, significantly reducing computational overhead.}
    \label{fig:method}
\end{figure}

We propose \algname, a framework transforming generative sampling into a closed-loop system by learning to integrate alignment feedback (Fig.~\ref{fig:method}). 
In the following sections, we 
investigate various formulations of the feedback signal and describe how the model is trained to internalize them.  Pseudocode for training and inference is provided in Algs.~\ref{alg:training} and \ref{alg:sampling} (Appendix).

\subsection{Problem Formulation}
\label{sec:problem_formulation}

Conditional generative modeling aims to sample from $p(\mathbf{x} \mid \mathbf{c})$, producing samples that satisfy two simultaneous objectives: \textbf{plausibility}, adhering to the target data manifold, and \textbf{fidelity}, maintaining consistency with the conditioning signal. Given a forward operator $\mathcal{H}$ accessible during training and inference, we define the observation $\mathbf{y} = \mathcal{H}(\mathbf{x})$. The condition $\mathbf{c}$ includes $\mathbf{y}$ and potentially auxiliary signals such as text. This formulation generalizes diverse tasks; for example, $\mathcal{H}$ may represent a renderer for 3D texturing, a depth predictor for depth-to-RGB, or a degradation operator for image restoration. In this context, the objective is to sample from the posterior distribution $p(\mathbf{x} \mid \mathbf{c}, \mathcal{H}).$

\subsection{The Two-Pass Feedback Loop}
\label{sec:two_pass_loop}
The core mechanism of \algname relies on feeding a task-specific feedback signal, derived from the model's current alignment error, back into the network as a first-class input. However, obtaining the clean signal estimate $\hat{\mathbf{x}}_1$ required to compute this feedback $\mathbf{s}_t$ introduces a causal dependency: $\mathbf{s}_t$ must be provided as a model input, yet it depends on $\hat{\mathbf{x}}_1$, which is only available from the model's output.
We resolve this via a \textbf{two-pass execution strategy} where a single network $\mathbf{v}_\theta$ handles both unguided and feedback-aware regimes. In the first pass, we set $\mathbf{s}_t = \mathbf{0}$ to generate a \emph{look-ahead} velocity $\mathbf{v}_{\text{LA}}$ and derive the initial estimate $\hat{\mathbf{x}}_1$. This estimate allows for computing deviations from target constraints to define the feedback signal $\mathbf{s}_t$. In the second pass, the model consumes $\mathbf{s}_t$ alongside $(\mathbf{x}_t, t, \mathbf{c})$ to produce the final, refined velocity $\mathbf{v}_{\text{ref}}$.

\subsection{Feedback Design Variants}
\label{sec:feedback_variants}

We propose several formulations for the error signal $\mathbf{s}_t$. A key design principle is to provide $\mathbf{s}_t$ as an auxiliary input naturally supported by any model, rendering our framework architecture-agnostic.

\vspace{-8pt}
\paragraph{First-Order Feedback.} 
For differentiable $\mathcal{H}$, we define an alignment loss $\mathcal{L}(\mathcal{H}(\mathbf{x}), \mathbf{y})$ quantifying the discrepancy between prediction and target. The first-order feedback signal is derived from the gradient of this loss. While Bayesian formulations (Eq.~\ref{eq:bayes_posterior}) require gradients w.r.t. $\mathbf{x}_t$, stability and efficiency often favor approximations w.r.t. the estimated clean signal $\hat{\mathbf{x}}_1$ \cite{he2023manifold, patel2025flowchef, kim2025flowdps}. We therefore evaluate two candidates for $\mathbf{s}^{\text{grad}}_t$: $\nabla_{\mathbf{x}_t} \mathcal{L}(\mathcal{H}(\hat{\mathbf{x}}_1), \mathbf{y})$ and the ``shortcut'' $\nabla_{\hat{\mathbf{x}}_1} \mathcal{L}(\mathcal{H}(\hat{\mathbf{x}}_1), \mathbf{y})$. The latter omits the expensive denoiser Jacobian $\frac{\partial \hat{\mathbf{x}}_1}{\partial \mathbf{x}_t}$, reducing memory overhead. The resulting gradient is concatenated to $\mathbf{x}_t$ along the channel dimension, providing an explicit direction for error correction.

\vspace{-8pt}
\paragraph{Zero-Order Feedback.} 
We additionally consider a derivative-free feedback signal defined by a measurement-space error operator:
$
\mathbf{s}^{\text{err}}_t = \mathcal{R}(\mathcal{H}(\hat{\mathbf{x}}_1), \mathbf{y}).
$
$\mathcal{R}$ is task-specific and aligns with the domain of the conditioning signal $\mathbf{y}$, $e.g.$, a pixel-wise residual for images. In this variant, the residual $\mathbf{s}^{\text{err}}_t$ is provided as a conditional input alongside $\mathbf{y}$, forcing the model to learn a mapping from measurement-space errors to signal-space updates. By avoiding gradients through $\mathcal{H}$, this approach improves efficiency and supports non-differentiable operators—such as JPEG compression or physical simulations—where gradient-based guidance is impossible. It also extends to black-box systems where gradients are unavailable, such as proprietary APIs or feature extractors.

\vspace{-8pt}
\paragraph{Hybrid Feedback.} 
Finally, we consider a composite variant that incorporates both first- and zero-order feedback. In this configuration, the model receives the gradient $\mathbf{s}^{\text{grad}}_t$ and the residual $\mathbf{s}^{\text{err}}_t$ through their respective input channels.

\subsection{Training}
\label{sec:training}

We optimize the model parameters $\theta$ using a joint training paradigm that supports both unguided and feedback-aware modes. For a pair $(\mathbf{x}_1, \mathbf{c})$ and timestep $t$, we generate $\mathbf{x}_t$ following Eq.~\ref{eq:forward_process}. First, the model computes a look-ahead estimate $\hat{\mathbf{x}}_1$ by setting the feedback input to zero ($\mathbf{s}_t = \mathbf{0}$). We then compute the feedback signal $\mathbf{s}_t$  from this estimate (Section~\ref{sec:feedback_variants}). The training objective is:
\[ \mathcal{L}_{\text{FA}} = \mathbb{E} [ \| \mathbf{v}_\theta(\mathbf{x}_t, t, \mathbf{c}, \text{sg}[\mathbf{s}_t]) - \mathbf{u}_t \|^2 ], \]
where $\text{sg}[\cdot]$ denotes the stop-gradient operation. To ensure numerical stability and efficiency, we treat $\mathbf{s}_t$ as a constant input rather than differentiating through the look-ahead pass and operator $\mathcal{H}$. To maintain look-ahead reliability, we randomly replace $\mathbf{s}_t$ with a null vector with probability $p_{\text{un}}$, similar to the training protocol of CFG \cite{ho2022classifier}. This joint training ensures the model remains accurate in the unguided regime, which is essential for generating reliable feedback at inference time.

\subsection{Inference}
\label{sec:inference}

Inference follows the two-pass procedure described in Section~\ref{sec:two_pass_loop}: at each step $t$, an unguided look-ahead pass estimates the clean signal $\hat{\mathbf{x}}_1$ to derive the feedback signal $\mathbf{s}_t$. A subsequent refinement pass then consumes $\mathbf{s}_t$ to yield the final velocity $\mathbf{v}_{\text{ref}}$ used to advance the ODE. We next describe optional modifications to this procedure that enable stronger alignment and more efficient sampling.

\begin{figure}[h!]
    \centering
    \includegraphics[width=1\linewidth]{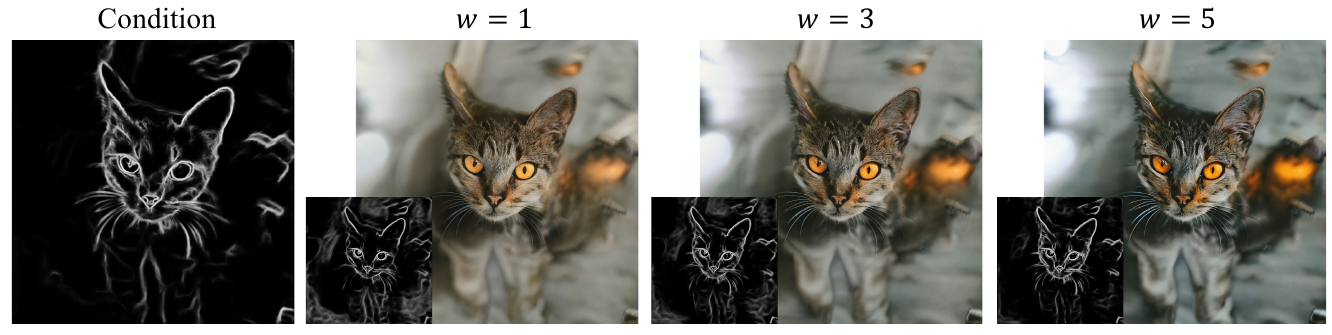}
    \caption{\textbf{Effect of Optional CFG.} Increasing guidance strength $w$ can enhance condition fidelity. Insets show re-extracted edge maps.}
    \label{fig:cfg_example}
\end{figure}

\vspace{-8pt}
\paragraph{Optional CFG.}
Our approach enables Classifier-Free Guidance (CFG) \cite{ho2022classifier} at zero marginal cost. While our framework aims to move beyond heuristic-based sampling, users desiring manual control can elegantly leverage the unguided velocity $\mathbf{v}_{\text{LA}}$, already computed to derive the feedback signal, as the ``unconditional'' reference. We define the resulting velocity as:
$
    \mathbf{v}_{\text{cfg}} = w \cdot \mathbf{v}_{\text{ref}} + (1-w) \cdot \mathbf{v}_{\text{LA}},
$ 
where $w$ modulates alignment strength (see Fig.~\ref{fig:cfg_example} and Appendix \ref{appendix:additional_results}).

\vspace{-10pt}
\paragraph{Efficient Sampling via Prior-Step Feedback Approximation.} The baseline two-pass execution doubles model evaluations as each step requires both a look-ahead and a refinement pass. To alleviate this, we propose an optional shortcut that exploits the similarity of error signals across subsequent timesteps to approximate the feedback without an additional evaluation. Analysis indicates that the feedback derived from the \emph{unguided} prediction at $t$ is increasingly correlated with the feedback computed from the \emph{guided} prediction of the preceding step as the trajectory approaches the clean manifold (see Fig. \ref{fig:t_thresh_ablation}(a--b)). This suggests that the refined prediction from the previous step serves as an effective surrogate for the current unguided clean estimate, particularly in the later stages of generation.  We introduce a threshold $t_{\text{thresh}} \in [0, 1]$ to control this approximation; for $t > t_{\text{thresh}}$, we bypass the look-ahead pass by deriving $\mathbf{s}_t$ from the cached estimate $\hat{\mathbf{x}}_1^{\text{prev}}$ from the prior step (Fig. \ref{fig:method}, bottom-right). This threshold enables a controllable trade-off between train-test compatibility and sampling speed. While $t_{\text{thresh}}=1$ preserves the two-pass logic, $t_{\text{thresh}}=0$ collapses execution into a single-pass loop after an initial bootstrap step -- requiring only $N+1$ evaluations for an $N$-step trajectory, nearly matching vanilla sampling efficiency while retaining closed-loop corrective benefits.

\section{Experiments}
\label{sec:experiments}

We evaluate \algname by fine-tuning pre-trained models for image-to-image translation and 3D mesh texturing. We conduct these evaluations using latent flow-matching models, as they represent the current state-of-the-art, though our method fits diffusion and signal-space models as well. We compare our closed-loop approach against three established paradigms for conditional sampling:

\begin{enumerate}[nosep, leftmargin=*]
\item \textbf{Standard Fine-Tuning (Standard FT):} The conventional single-pass supervised approach for learning a conditional velocity field. We implement this using ControlNet \cite{zhang2023adding} and LoRA \cite{hu2022lora}, two popular adapters designed to incorporate new conditions into pre-trained models while mitigating catastrophic forgetting.
\item \textbf{Fine-Tuning with Alignment Loss (FT + $\mathcal{L}_{\text{align}}$):} An augmented training strategy that incorporates an explicit measurement-consistency objective during the fine-tuning stage to improve fidelity, as proposed in ControlNet++ \cite{li2024controlnet++}.
\item \textbf{Inference-Time Guidance (IT Guidance):} A test-time approach that applies gradient updates during sampling to enforce conditional consistency without retraining. Specifically, we compare against FlowChef \cite{patel2025flowchef}, a recent state-of-the-art general guidance framework for FM models, as the representative baseline for this paradigm.
\end{enumerate}

\vspace{-10pt}
\paragraph{Experimental Protocol.} Performance is evaluated across two primary axes: \emph{fidelity}, defined by the alignment between the projected output $\mathcal{H}(\mathbf{x})$ and the target measurement $\mathbf{y}$, and \emph{plausibility}, which assesses adherence to the target data manifold (e.g., perceptual quality). To ensure a controlled comparison and isolate the impact of our feedback mechanism, primary evaluations utilize an identical number of sampling steps across all methods. We provide more extensive comparisons in Appendix \ref{appendix:additional_results}, including results for baselines with doubled sampling budgets and the incorporation of CFG. As we demonstrate, such standard test-time enhancements do not resolve the fundamental shortcomings of existing paradigms.

\subsection{Image-to-Image Translation}
\label{sec:exp_translation}

We evaluate our framework on four image-to-image translation tasks: super-resolution, depth-to-RGB, edge-to-RGB, and JPEG restoration. These represent closed-form, neural-network-based, and non-differentiable forward operators $\mathcal{H}$, respectively. Our setup uses Stable Diffusion 3.5 Large~\cite{esser2024scaling} with ControlNet~\cite{zhang2023adding} for conditioning. We train on the Unsplash-25K~\cite{ali2023unsplash} dataset (20k training and 5k test images) and sample using the Euler sampler with 40 steps.
The forward operators consist of an 8$\times$ downsampling kernel (SR), DepthAnythingV2~\cite{yangDepthAnythingV22024} (depth), HED~\cite{xie2015hed} (edges), and JPEG compression ($\sigma=10$). To assess \emph{fidelity} compared to the provided condition image, we report PSNR, SSIM, and LPIPS for restoration tasks; MAE and MSE for edges; and MAE and $\delta_{1.25}$ for depth. \emph{Plausibility} is measured using FID across all tasks.
Since JPEG restoration is non-differentiable, we only compare Standard FT and our zero-order variant for this task, as other baselines and closed-loop variants require a differentiable operator. 

\cref{tab:SR_depth_edge_generation,tab:jpeg_restoration} summarize the quantitative results across all four tasks. \algname's variants consistently outperform the baselines, yielding significant gains in both fidelity and plausibility. \textit{IT Guidance} exhibits a strict trade-off between fidelity and plausibility depending on hyperparameter choices. While competitive on fidelity metrics, we find that this is traded off with worse quality of generated images as indicated by higher FID and vice versa.
We provide further results and discussion on this trade-off in Appendix \ref{appendix:image_to_image}. Qualitative examples for edge and depth to RGB are provided in \cref{fig:image_to_image}, with restoration tasks examples available in the Appendix.

\begin{table*}[t]
\centering
\caption{\textbf{Image-to-Image Results.} \colorbox{gray!10}{Shaded rows} denote \algname variants. Best results in \textbf{bold}; second best \underline{underlined}.}
\label{tab:SR_depth_edge_generation}
\adjustbox{max width=\linewidth}{%
\begin{tabular}{l ccc c cc c cc c}
  \toprule
  & \multicolumn{4}{c}{\textbf{Super Resolution}} & \multicolumn{3}{c}{\textbf{Depth-to-RGB}} & \multicolumn{3}{c}{\textbf{Edge-to-RGB}} \\
  \cmidrule(lr){2-5} \cmidrule(lr){6-8} \cmidrule(lr){9-11}
  & \multicolumn{3}{c}{Fidelity} & Plausibility & \multicolumn{2}{c}{Fidelity} & Plausibility & \multicolumn{2}{c}{Fidelity} & Plausibility \\
  \cmidrule(lr){2-4} \cmidrule(lr){5-5} \cmidrule(lr){6-7} \cmidrule(lr){8-8} \cmidrule(lr){9-10} \cmidrule(lr){11-11}
  Method & PSNR $\uparrow$ & SSIM $\uparrow$ & LPIPS $\downarrow$ & FID $\downarrow$ & MAE $\downarrow$ & $\delta_{1.25} \uparrow$ & FID $\downarrow$ & MAE $\downarrow$ & MSE $\downarrow$ & FID $\downarrow$ \\
  \midrule
  Standard FT & 34.35 & 96.88 & 0.83 & 3.93 &  0.0848 & 0.7882 & 18.21 & 0.0533 & 0.0137 & 13.98 \\
  FT + $\mathcal{L}_{\text{align}}$ & 35.21 & 97.53 & 0.79 & 4.11 & 0.0834 & 0.7933 & 17.57 & 0.0501 & 0.0128 & 14.47 \\
  IT Guidance & \underline{43.02} & \underline{98.29} & 0.65 & 18.96 & 0.0866 & 0.7592 & 223.54 & \underline{0.0416} & 0.0129 & 97.65 \\
  \midrule
  \rowcolor{gray!10} First-order (w.r.t. $\mathbf{x}_t$) & 36.27 & 97.10 & 0.66 & 4.30 & \textbf{0.0747} & \textbf{0.8268} & \textbf{14.57} & 0.0456 & 0.0104 & 14.81 \\
  \rowcolor{gray!10} First-order (w.r.t. $\hat{\mathbf{x}}_1$) & \textbf{44.07} & \textbf{98.96} & 0.33 & 6.45 & 0.0818 & 0.7973 & 15.89 &  0.0435 & 0.0123 & 14.97 \\
  \rowcolor{gray!10} Zero-order & 39.25 & 98.18 & \textbf{0.21} & \textbf{3.36} & \underline{0.0764} & \underline{0.8187} & 15.70 & 0.0460 & 0.0111 & 14.29 \\
  \rowcolor{gray!10} Combined (w.r.t. $\mathbf{x}_t$) & 39.95 & 98.25 & \textbf{0.21} & 3.40 & 0.0829 & 0.7949 & 15.93 & 0.0435 & \underline{0.0097} & \underline{13.91} \\
  \rowcolor{gray!10} Combined (w.r.t. $\hat{\mathbf{x}}_1$) & 39.77 & 98.24 & \underline{0.23} & \underline{3.39} & 0.0783 & 0.8175 & \underline{15.39} & \textbf{0.0408} & \textbf{0.0085} & \textbf{13.68} \\
  \bottomrule
\end{tabular}
}
\end{table*}

\begin{table*}[t]
\centering
\begin{minipage}{0.6\textwidth}
    \centering
    \caption{\textbf{JPEG Restoration Results.} }
    \label{tab:jpeg_restoration}
    \adjustbox{max width=\linewidth}{%
    \begin{tabular}{l ccc c}
        \toprule
        & \multicolumn{3}{c}{Fidelity} & Plausibility \\
        \cmidrule(lr){2-4} \cmidrule(lr){5-5}
        Method & PSNR $\uparrow$ & SSIM $\uparrow$ & LPIPS $\downarrow$ & FID $\downarrow$ \\
        \midrule
        Standard FT & 26.29 & 79.45 & 22.24 & 4.35 \\
        \rowcolor{gray!10} Zero-order (Ours) & \textbf{28.86} & \textbf{83.13} & \textbf{16.33} & \textbf{3.80} \\
        \bottomrule
    \end{tabular}}
\end{minipage}
\hfill
\begin{minipage}{0.35\textwidth}
    \centering
    \caption{\textbf{Ablation of $p_{un}$.}}
    \label{tab:ablation_p_un}
    \adjustbox{max width=\linewidth}{%
    \begin{tabular}{lcccc}
        \toprule
         & \multicolumn{3}{c}{Fidelity} & Plausibility \\
        \cmidrule(lr){2-4} \cmidrule(lr){5-5}
        $p_{un}$ & PSNR $\uparrow$ & SSIM $\uparrow$ & LPIPS $\downarrow$ & FID $\downarrow$ \\
        \midrule
        0.0 & 37.38 & 97.41 & 0.49 & 3.89 \\
        0.1 & \textbf{39.21} & \textbf{97.64} & \textbf{0.43} & \textbf{3.83} \\
        0.2 & 38.55 & 97.57 & 0.48 & 3.93 \\
        0.3 & 37.55 & 97.24 & 0.53 & 3.92 \\
        \bottomrule
    \end{tabular}}
\end{minipage}
\end{table*}

\begin{figure}[t]
    \centering
    \includegraphics[width=1\linewidth]{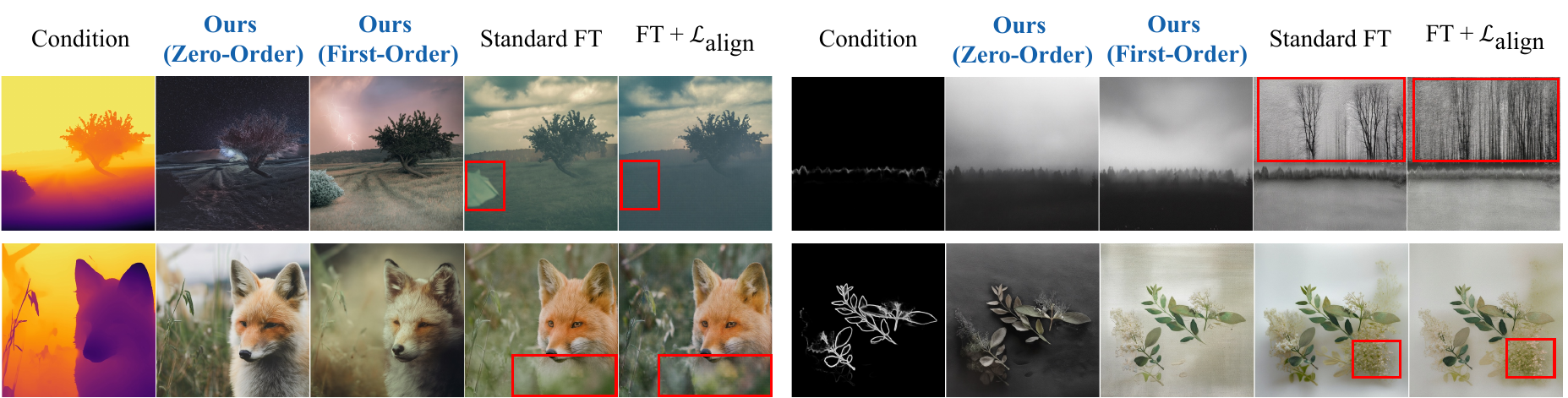}
    \caption{\textbf{Qualitative comparisons.} (Left) Depth-to-RGB; (right) Edge-to-RGB. Red boxes highlight conditioning inconsistencies.} 
    \label{fig:image_to_image}
\end{figure}

\subsection{3D Mesh Texturing}
\label{sec:exp_3d}
We evaluate \algname on 3D mesh texturing by fine-tuning the TRELLIS-2 \cite{xiang2025native} texture transformer. Given the 3D geometry and the conditioning image $\mathbf{y}$ as inputs, we utilize their corresponding latents as conditions, integrate LoRA adapters \cite{hu2022lora} into all linear layers and expand the input channels to accommodate the feedback signal $\mathbf{s}_t$. We focus on first-order feedback, concatenating $\mathbf{s}_t$ with noisy latents, avoiding the complexity of injecting zero-order residuals via DINO-based cross-attention \cite{simeoni2025dinov3}. The forward operator $\mathcal{H}$ is defined as the composition of the TRELLIS-2 latent decoder and its differentiable PBR renderer. Training uses 7500 Objaverse assets \cite{deitke2023objaverse}, with evaluation on 100 held-out Objaverse and 100 Toys4K assets \cite{stojanov2021using}.
Fidelity is measured via masked PSNR (M.PSNR, excluding background), SSIM, LPIPS, and CLIP-similarity to the conditioning image. Plausibility is assessed by evaluating 50 random renders for each asset (5,000 total images). From these, we compute FID and standard multi-view reconstruction metrics. We adopt the default TRELLIS-2 configuration ($n=12$ steps, Euler sampler). 

\begin{figure}
    \centering
    \includegraphics[width=1\linewidth]{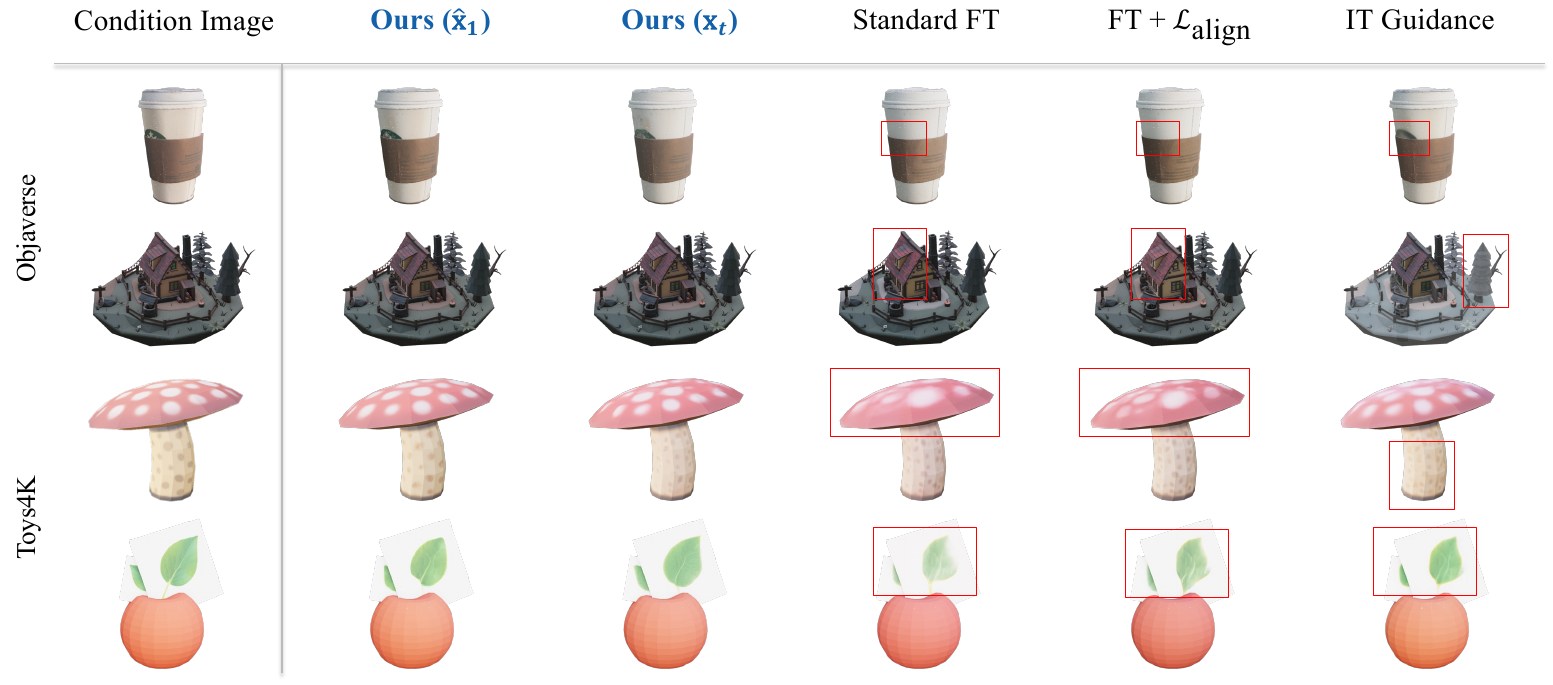}
    \caption{\textbf{3D Texturing Results.} Objaverse (rows 1--2) and Toys4K (3--4) assets. The leftmost column provides the input condition image; remaining columns show generated 3D textured assets rendered from corresponding viewpoints. Red boxes highlight conditioning inconsistencies.}
    \label{fig:texturing_comp}
\end{figure}

Quantitative results in Table \ref{tab:3d_results_combined} show our framework consistently outperforms all baselines. Notably, the $\nabla_{\hat{\mathbf{x}}_1}$ variant achieves the strongest performance, significantly improving over supervised baselines. While IT Guidance enhances fidelity, it fails to match \algname in plausibility. Qualitative comparisons (Fig. \ref{fig:texturing_comp}) illustrate that our approach recovers fine-grained details omitted by baselines. Additional visualizations, including multi-view renderings, are provided in the Appendix.

\begin{table*}[t]
\centering
\caption{\textbf{Mesh Texturing Results.} Metrics evaluate single-view fidelity to the conditioning image and multi-view (MV) plausibility of the resulting 3D texture across Objaverse and Toys4K datasets. \colorbox{gray!10}{Shaded rows} denote \algname variants. Best results in \textbf{bold}; second best \underline{underlined}.}
\label{tab:3d_results_combined}
\adjustbox{max width=\linewidth}{%
\begin{tabular}{l cccc ccccc}
\toprule
& \multicolumn{4}{c}{\textbf{Fidelity}} & \multicolumn{5}{c}{\textbf{Plausibility}} \\
\cmidrule(lr){2-5} \cmidrule(lr){6-10}
Method & M. PSNR $\uparrow$ & SSIM $\uparrow$ & LPIPS $\downarrow$ & CLIP $\uparrow$ & MV-M.PSNR $\uparrow$ & MV-SSIM $\uparrow$ & MV-LPIPS $\downarrow$ & MV-CLIP $\uparrow$ & FID $\downarrow$ \\
\midrule
\multicolumn{10}{l}{\textit{Objaverse Dataset}} \\
\midrule
Pre-Trained &  19.20 & 0.978 & 0.0221 & 0.961 & 19.91 & 0.987 & 0.0138 & 0.961 & 11.03 \\
IT Guidance &  \underline{25.86} & \underline{0.986} & 0.0191 & 0.973 & 22.80 & 0.989 & 0.0123 & 0.968 & 9.10 \\
Standard FT & 21.91 & 0.981 & 0.0189 & 0.969 & 23.05 & 0.990 & 0.0115 & 0.970 & 8.74 \\
FT + $\mathcal{L}_{\text{align}}$ & 22.63 & 0.983 & 0.0182 & 0.970 & 22.48 & 0.989 & 0.0117 & 0.969 & 8.94 \\
\midrule

\rowcolor{gray!10} Ours (w.r.t. $\mathbf{x}_t$) & 25.26 & 0.985 & \underline{0.0157} & \underline{0.979} & \underline{25.91} & \underline{0.991} & \underline{0.0105} & \underline{0.974} & \underline{7.41} \\
\rowcolor{gray!10} Ours (w.r.t. $\hat{\mathbf{x}}_1$) & \textbf{26.39} & \textbf{0.987} & \textbf{0.0140} & \textbf{0.983} & \textbf{26.53} & \textbf{0.992} & \textbf{0.0098} & \textbf{0.977} & \textbf{6.64} \\
\midrule
\multicolumn{10}{l}{\textit{Toys4K Dataset}} \\
\midrule
Pre-Trained &  20.36 & 0.984 & 0.0182 & 0.965 & 20.63 & 0.989 & 0.0129 & 0.966 & 10.66 \\
IT Guidance &  \underline{26.82} & \underline{0.990} & 0.0137 & \underline{0.982} & 23.63 & 0.992 & 0.0112 & 0.972 & 8.43 \\
Standard FT & 23.00 & 0.986 & 0.0160 & 0.975 & 24.38 & 0.991 & 0.0114 & 0.972 & 8.87 \\
FT + $\mathcal{L}_{\text{align}}$ & 23.05 & 0.986 & 0.0159 & 0.976 & 23.25 & 0.991 & 0.0111 & 0.973 & 8.30 \\
\midrule
\rowcolor{gray!10} Ours (w.r.t. $\mathbf{x}_t$) & 26.54 & \underline{0.990} & \underline{0.0129} & 0.980 & \underline{27.20} & \underline{0.993} & \underline{0.0097} & \underline{0.977} & \underline{7.20} \\
\rowcolor{gray!10} Ours (w.r.t. $\hat{\mathbf{x}}_1$) & \textbf{27.26} & \textbf{0.991} & \textbf{0.0116} & \textbf{0.984} & \textbf{27.21} & \textbf{0.993} & \textbf{0.0092} & \textbf{0.978} & \textbf{6.66} \\
\bottomrule
\end{tabular}
}
\end{table*}

\begin{figure}[t]
    \centering
    \includegraphics[width=\linewidth]{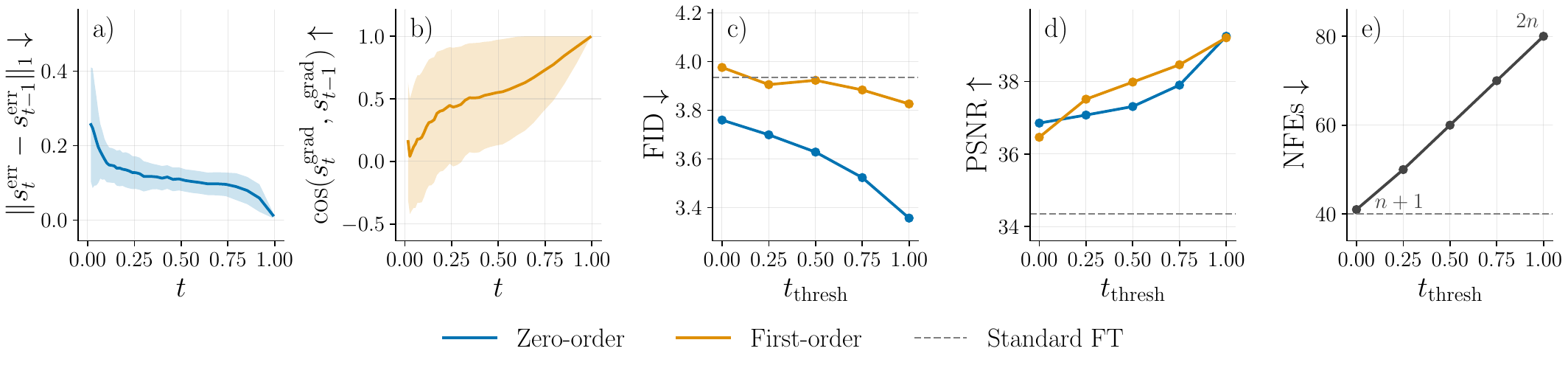}
\caption{\textbf{Prior-Step Shortcut Analysis.} (a--b) Temporal similarity of feedback signals for zero-order (a) and first-order (b) variants; rising correlation as $t \to 1$ motivates the $t_{\text{thresh}}$-controlled shortcut strategy. (c--d) FID and PSNR vs.\ $t_{\text{thresh}}$; our method maintains a consistent advantage over Standard FT (dashed) even at low $t_{\text{thresh}}$ values. (e) Computational cost (NFEs) as a function of $t_{\text{thresh}}$, where $n$ is the number of sampling steps.}
    \label{fig:t_thresh_ablation}
\end{figure}

\subsection{Ablation Study}
\label{sec:ablation}
We evaluate the impact of key hyperparameters on the super-resolution task.
\vspace{-10pt}
\paragraph{Inference-time Shortcut.}We analyze the efficiency-performance trade-off controlled by $t_{\text{thresh}}$ (Section \ref{sec:inference}). As shown in Fig. \ref{fig:t_thresh_ablation}, reducing $t_{\text{thresh}}$ decreases the computational cost toward $n+1$ NFEs, approaching the speed of vanilla sampling at lower threshold values. Although the shortcut introduces a relative performance dip, our approach consistently outperforms the Standard FT baseline even at low $t_{\text{thresh}}$ settings. The fidelity gap at $t_{\text{thresh}}=0$ is especially notable. These results demonstrate that while our primary approach utilizes a two-pass logic, the framework can provide substantial corrective benefits even with minimal computational overhead compared to standard open-loop sampling.

\vspace{-10pt}
\paragraph{Null Feedback Probability.}
The parameter $p_{\text{un}}$ allocates the training budget between unguided and feedback-aware iterations. Conceptually, a non-zero $p_{\text{un}}$ supports the model's ability to generate reliable look-ahead estimates, which serve as the foundation for the feedback signal. However, increasing this probability limits the iterations available for learning the second-pass refinement. Table \ref{tab:ablation_p_un} shows that $p_{un}=0.1$ yields the best results across all metrics, providing the optimal balance for the closed-loop system.

\subsection{Is \algname Just Gradient Guidance in Disguise?}
\label{sec:analysis}
We investigate whether closed-loop training simply automates hyperparameter tuning for linear guidance. Evaluating 180 velocity predictions across 20 assets in the texturing task ($\nabla_{\hat{\mathbf{x}}_1}$ variant), we decompose the learned correction $\Delta \mathbf{v} = \mathbf{v}_{\text{ref}} - \mathbf{v}_{\text{LA}}$ relative to the gradient signal $\mathbf{s}_t^{\text{grad}}$. The squared norm of the parallel component accounts for approximately 20\% of the total correction's energy; the majority of the correcting update resides in the component orthogonal to the gradient. This relationship remains stable across noise levels ($t \in [0.1, 0.9]$). While a cosine similarity of $\cos(\Delta \mathbf{v}, \mathbf{s}_t^{\text{grad}}) = 0.42 \pm 0.11$ confirms the gradient is utilized, the dominant orthogonal component proves that \algname internalizes the gradient as a high-dimensional feature to non-linearly \emph{bend} the flow toward the manifold, transcending the strict additive formulation of traditional Bayesian guidance.

\section{Conclusion and Limitations}
\label{sec:conclusion}

We presented \algname, a framework that addresses the fundamental "open-loop" limitation of existing conditional flow models by treating their own alignment errors as first-class inputs, thus training them to correct (``bend'') their initial predictions. By replacing hand-tuned guidance with a learned,  non-linear two-pass correction policy, \algname simultaneously enhances both conditional fidelity and sample plausibility. Our approach significantly improves over supervised and inference-time baselines across a diverse range of tasks, including image-to-image translation, restoration, and 3D mesh texturing.

Despite these gains, certain limitations remain. First, while our prior-step shortcut enables efficient inference, the training phase requires an additional model evaluation per iteration to derive the feedback signal, increasing the computational budget for fine-tuning. Future work could investigate training schemes that directly utilize cached prior-step predictions, potentially restoring single-pass efficiency throughout the entire pipeline. Second, performance sometimes further improves when using CFG alongside \algname (Sec.~\ref{sec:inference}), suggesting the learned policy has not yet fully internalized the most complex conditioning nuances. We believe that more expressive feedback-integration architectures and large-scale training are promising directions for closing this gap.

\section{Acknowledgments}
Or Litany acknowledges support from the Israel Science Foundation (grant 624/25) and the Azrieli Foundation Early Career Faculty Fellowship. This research was also supported in part by an academic gift from Meta. The authors gratefully acknowledge this support. This research was supported by the Council for Higher Education in Israel under the Moonshot Project.

\bibliographystyle{plainnat}
\bibliography{references}

\clearpage

\appendix

\section{Implementation Details}
\label{appendix:implementation_details}

\begin{algorithm}[H]
    \refstepcounter{algorithm} \label{alg:training}
    \centering
    {\textsc{Algorithm \thealgorithm: Feedback-Aware Training}}
    \vspace{2pt} \hrule height 0.4pt \vspace{4pt}
    \small
    \begin{algorithmic}[1]
        \Require Dataset $\mathcal{D}$, model $\mathbf{v}_\theta$, prob. $p_{\text{un}}$
        \While{not converged}
            \State Sample $(\mathbf{x}_1, \mathbf{c}) \sim \mathcal{D}$ where $\mathbf{y} \in \mathbf{c}$
            \State $\mathbf{x}_0 \sim \mathcal{N}(0, \mathbf{I}), t \sim \mathcal{U}[0, 1]$ 
            \State $\mathbf{x}_t \gets a_t \mathbf{x}_1 + \sigma_t \mathbf{x}_0$
            \State $\mathbf{v}_{\text{LA}} \gets \mathbf{v}_\theta(\mathbf{x}_t, t, \mathbf{c}, \mathbf{0})$ \hfill \textit{\textcolor{algcomment}{// Pass 1: Unguided Look-ahead}}
            \State $\hat{\mathbf{x}}_1 \gets \operatorname{EstimateSample}(\mathbf{x}_t, t, \mathbf{v}_{\text{LA}})$ 
            \State $\mathbf{s}_t \gets \operatorname{Feedback}(\hat{\mathbf{x}}_1, \mathbf{y})$ \hfill \textit{\textcolor{algcomment}{// See Sec. \ref{sec:feedback_variants}}}
            \If{$\operatorname{rand}(0,1) < p_{\text{un}}$} \hfill \textit{\textcolor{algcomment}{// Conditioning Dropout}}
                \State $\mathbf{s}_{\text{in}} \gets \mathbf{0}$
            \Else
                \State $\mathbf{s}_{\text{in}} \gets \operatorname{stop\_grad}(\mathbf{s}_t)$ 
            \EndIf
            \State $\mathbf{v}_{\text{ref}} \gets \mathbf{v}_\theta(\mathbf{x}_t, t, \mathbf{c}, \mathbf{s}_{\text{in}})$ \hfill \textit{\textcolor{algcomment}{// Pass 2: Refinement}}
            \State $\mathcal{L}_{\text{FA}} \gets \| \mathbf{v}_{\text{ref}} - \mathbf{u}_t \|^2$ 
            \State Update $\bm{\theta}$ via $\nabla_{\bm{\theta}} \mathcal{L}_{\text{FA}}$
        \EndWhile
    \end{algorithmic}
    \vspace{2pt} 
\end{algorithm}

\begin{algorithm}[H]
    \refstepcounter{algorithm} \label{alg:sampling}
    \centering
    {\textsc{Algorithm \thealgorithm: Feedback-Aware Inference}}
    \vspace{2pt} \hrule height 0.4pt \vspace{4pt}
    \small
    \begin{algorithmic}[1]
        \Require $\mathbf{v}_\theta$, condition $\mathbf{c}$ (incl. $\mathbf{y}$), $t_{\text{thresh}}$, ODE Solver
        \State $\mathbf{x}_0 \sim \mathcal{N}(0, \mathbf{I})$ 
        \State $\hat{\mathbf{x}}_1^{\text{prev}} \gets \mathbf{0}$
        \For{$i = 0$ to $N-1$}
            \State $t \gets t_i$
            \If{$t \leq t_{\text{thresh}}$ \textbf{or} $i=0$} \hfill \textit{\textcolor{algcomment}{// Pass 1: Unguided Look-ahead}}
                \State $\mathbf{v}_{\text{LA}} \gets \mathbf{v}_\theta(\mathbf{x}_t, t, \mathbf{c}, \mathbf{0})$
                \State $\hat{\mathbf{x}}_1 \gets \operatorname{EstimateSample}(\mathbf{x}_t, t, \mathbf{v}_{\text{LA}})$
            \Else \hfill \textit{\textcolor{algcomment}{// Shortcut: Prior-Step Reuse}}
                \State $\hat{\mathbf{x}}_1 \gets \hat{\mathbf{x}}_1^{\text{prev}}$
            \EndIf
            \State $\mathbf{s}_t \gets \operatorname{Feedback}(\hat{\mathbf{x}}_1, \mathbf{y})$
            \State $\mathbf{v}_{\text{ref}} \gets \mathbf{v}_\theta(\mathbf{x}_t, t, \mathbf{c}, \mathbf{s}_t)$ \hfill \textit{\textcolor{algcomment}{// Pass 2: Refinement}}
            \State $\hat{\mathbf{x}}_1^{\text{prev}} \gets \operatorname{EstimateSample}(\mathbf{x}_{t}, t_i, \mathbf{v}_{\text{ref}})$ \hfill \textit{\textcolor{algcomment}{// Cache Update}}
            \State $\mathbf{x}_{t+1} \gets \operatorname{Step}(\mathbf{x}_t, \mathbf{v}_{\text{ref}}, t)$ \hfill \textit{\textcolor{algcomment}{// Integration}}
        \EndFor
        \State \Return $\mathbf{x}_N$
    \end{algorithmic}
    \vspace{2pt} 
\end{algorithm}

\subsection{2D Toy Experiment}
\label{appendix:2d_toy}
\textbf{Ground-Truth Distribution.} The target distribution is a 2D Archimedean spiral with finite thickness, defined by the curve

$$r = a\theta, \quad \theta \sim \mathrm{Uniform}(0, 2\pi), \quad a = 2.0$$

The finite thickness of is
modeled by an isotropic Gaussian spread around each point on the curve ($\sigma = 0.12$).

Each point is assigned a quadrant label $c \in \{0,1,2,3\}$ based on which quarter-turn of the spiral it falls in ($c = \lfloor \theta / (\pi/2) \rfloor$).

\textbf{Model Architecture.} The generative model is an MLP with 3 linear layers, hidden dimension 64, and SiLU activations between layers. The input is a concatenation of the noisy sample $x_t \in \mathbb{R}^2$ with a time embedding. The scalar $t \in [0,1]$ is lifted to a 16-dimensional sinusoidal embedding~\cite{vaswani2017attention}. The class condition is encoded via a learned embedding table that maps the quadrant label $c$ to a vector of the hidden dimension, which is added to the time embedding before being passed to the main network.

\textbf{Training.}
All models are trained with the standard conditional flow matching
objective~\cite{lipman2022flow} for 50{,}000 steps with a batch size of 1024 and a learning rate of $3 \times 10^{-4}$ (AdamW optimizer). Training samples are generated on the fly at each iteration. Prior to training, a z-score normalizer is fitted on 200{,}000 spiral samples (per-coordinate mean and standard deviation). All inputs and model outputs are normalized using this transformation.

\textbf{Feedback and Guidance objective}
The spiral is divided into four quadrants, each assigned a non-overlapping arc-length band. For quadrant $c$, the target radius interval is:

$$r^c_{\min} = c\pi, \qquad r^c_{\max} = (c+1)\pi$$

The loss penalises points outside this band:

$$\mathcal{L}(x, c) = \text{softplus}(r^c_{\min} - r) + \text{softplus}(r - r^c_{\max}), \quad r = |x|_2$$

This is zero (up to softplus smoothing) when $r \in [r^c_{\min}, r^c_{\max}]$, and grows linearly outside.

\textbf{Guidance Scale.} As illustrated in Fig. \ref{fig:toy_exp}, Training-Free Guidance tends to push the final samples off the data manifold. We explore different guidance scales in~\cref{appendix:fig:2d_guidance_scales}. Small guidance scales fail to enforce the conditioning signal, while large guidance scales lead to samples that deviate from the target manifold.

\begin{figure}[t]
    \centering
    \includegraphics[width=\linewidth]{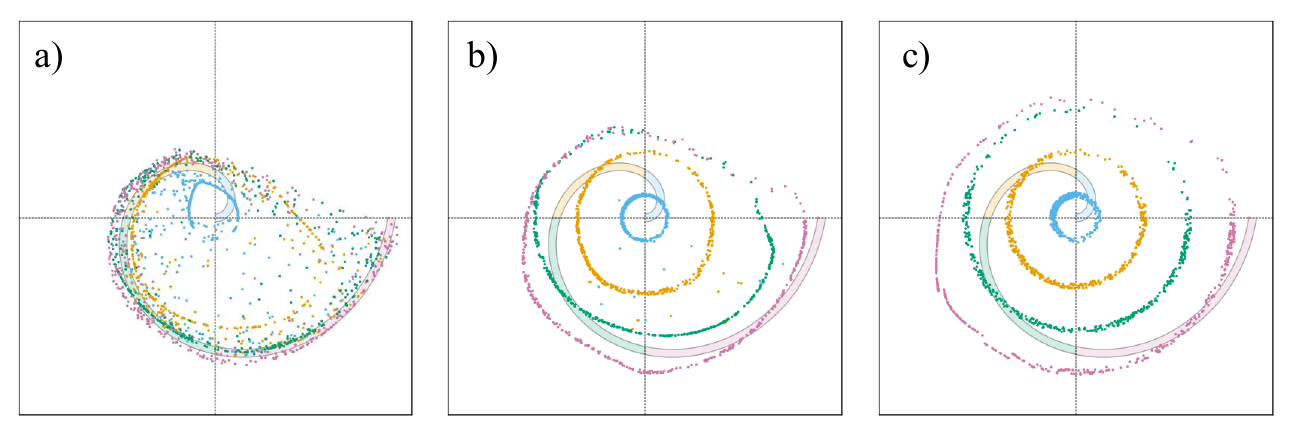}
    \caption{\textbf{Effect of guidance scale on conditional generation.} We compare guidance scales of a) 0.5, b) 2.0 (used in Fig. \ref{fig:toy_exp}), and c) 4.0. Small scale (a) fails to enforce the condition, while large guidance (c) pushes samples off the data manifold entirely. We therefore selected an intermediate scale that provides a reasonable trade-off between fidelity and plausibility.}
    \label{appendix:fig:2d_guidance_scales}
\end{figure}

\subsection{Image-to-Image Translation}

We use the official Stable Diffusion 3.5 (SD3.5) checkpoint implemented in the diffusers~\cite{von-platen-etal-2022-diffusers} library and adapt their reference script to add and train the ControlNets \footnote{\href{https://github.com/huggingface/diffusers/blob/main/examples/controlnet/train_controlnet_sd3.py}{https://github.com/huggingface/diffusers/blob/main/examples/controlnet/train\_controlnet\_sd3.py}}.

The model is  conditioned on $\mathbf{c} = (\mathbf{y}, \mathbf{c}_\mathrm{text})$ where $\mathbf{y}$ is the image-shaped condition (e.g., edges) and $\mathbf{c}_\mathrm{text}$ is the text condition. In the following, we omit the dependence on the text condition $\textbf{c}_\mathrm{text}$ for clarity.
\paragraph{Zero-order.}
To inject zero-order information, we expand the input convolution of the ControlNet and concatenate the residual to the condition along the channel dimension. Specifically, the ControlNet module $\operatorname{CN}$ computes the condition in the unguided pass from the image condition $\mathbf{y}$ as
$$
\mathbf{\hat{c}} = \operatorname{CN}([\mathbf{y}, \mathbf{0}]), \quad \mathbf{y}, \mathbf{0} \in \mathbb{R}^{3\times H\times W}
$$
where $[\cdot]$ denotes concatenation along the channel dimension.

We then compute the residual via 
\begin{align*}
\mathbf{v}_{\text{LA}} &= \mathbf{v}_\theta(\mathbf{x}_t, t, \mathbf{\hat{c}}) \\
\hat{\mathbf{x}}_1 &= \operatorname{EstimateSample}(\mathbf{x}_t, t, \mathbf{v}_{\text{LA}}) \\
\mathbf{\tilde{s}}_t &= \left|\mathbf{y} - \mathcal{H}(\hat{\mathbf{x}}_1)\right|
\end{align*}
where $\mathcal{H}$ is the composition of the VAE decoder and the pixel-space forward operator.
We normalize the feedback signal using 
$$
\mathbf{s}_t = \mathbf{\tilde{s}}_t / \max \left(| \mathbf{\tilde{s}}_t|\right)
$$
Finally, the ControlNet encodes the guidance signal $\mathbf{\tilde{c}} = \operatorname{CN}([\mathbf{y}, \mathbf{s}_t])$ and the refined velocity is 
$$
\mathbf{v}_{\text{ref}} = \mathbf{v}_\theta(\mathbf{x}_t, t, \mathbf{\tilde{c}}).
$$

\paragraph{First-order.}
For our first-order variant, we double the number of input channels of the input convolution of the DiT to inject the feedback signal. In the unguided pass, the denoising network $\mathbf{v}_\theta$ predicts
$$
\mathbf{v}_{\text{LA}} = \mathbf{v}_\theta \left([\mathbf{x}_t, \mathbf{0}], t, \mathbf{\tilde{c}}\right), \quad \mathbf{x}_t, \mathbf{0} \in \mathbb{R}^{16\times h\times w} 
$$
where $h, w$ are the height and width of the VAE-encoded image latents and $\mathbf{\tilde{c}} = \operatorname{CN}(\mathbf{y})$.

We then compute the guidance signal $\mathbf{\tilde{s}}_t = \nabla_{\mathbf{x}_t} \mathcal{L}$ where
\begin{align*}
\hat{\mathbf{x}}_1 &= \operatorname{EstimateSample}(\mathbf{x}_t, t, \mathbf{v}_{\text{LA}}) \\
\mathcal{L} &= \mathrm{MSE}(\mathbf{y}, \mathcal{H}(\hat{\mathbf{x}}_1))) \\
\end{align*}

We standardize the feedback signal 
$$
\mathbf{s}_t = \left( \mathbf{\tilde{s}}_t - \operatorname{mean}(\mathbf{\tilde{s}}_t) \right) / \operatorname{std}(\mathbf{\tilde{s}}_t)
$$
before computing the refined velocity as
$$
\mathbf{v}_{\text{ref}} = \mathbf{v}_\theta([\mathbf{x}_t, \mathbf{s}_t], t, \mathbf{\tilde{c}}).
$$
Inspired by \citet{patel2025flowchef}, we additionally explore a variant that uses $\mathbf{\tilde{s}}_t = \nabla_{\hat{\mathbf{x}}_1} \mathcal{L}$ which avoids differentiating through the denoising network $\mathbf{v}_\theta$.

\paragraph{Training}
We center crop and resize images of the Unsplash-25K~\cite{ali2023unsplash} dataset to $1,024^2$ resolution and create text conditions for images using Florence2-Large~\cite{xiao2024florence}.
Training uses batch size 16 for 5 epochs, resulting in a total of $6,250$ optimizer steps, using the AdamW~\cite{loshchilovDecoupledWeightDecay2019} optimizer with learning rate $10^{-5}$, weight decay $0.01$, and 500 steps of learning rate warmup.
Similar to \citet{zhang2023adding}, the weights corresponding to the additional channels, used to inject feedback information, are zero-initialized such that the network is slowly adapted to make use of feedback during training. We drop the feedback condition during training with probability $p_{un} = 0.1$ following our ablation study in the main text (Tab. \ref{tab:ablation_p_un}).

We only train newly added parameters, specifically the ControlNet, while keeping weights of the StableDiffusion3.5 base model frozen. In the first-order and combined variants, we additionally unfreeze the input convolution to the DiT~\cite{peebles2023scalable} to allow the model to make use of the feedback signal that is concatenated to the input latent. 

All methods use NVIDIA A100 GPUs for training where zero-order experiments take $\approx 51$ GPU hours, first-order experiments take $\approx 61$ GPU hours, and Standard FT and FT+$\mathcal{L}_{\text{align}}$ runs take $\approx 38$ and $\approx 48$ GPU hours, respectively.

\paragraph{Inference.}
We use 40 Euler steps with full two-pass execution at every step ($t_{\text{thresh}} = 1$). No CFG is applied in the main text comparisons (guidance scale 1.0), consistent with all other evaluated methods.

\paragraph{Baselines.} 

\textit{Standard FT.} Baseline fine-tuning variants use the vanilla ControlNet setup with zero-convolutions to slowly expose the network to the condition during training. 

\textit{FT + $\mathcal{L}_{\text{align}}$.} The variants with $\mathcal{L}_{\text{align}}$ adopt the set up in \citet{li2024controlnet++} with loss weight $\lambda_{\text{align}}=0.5$ for the consistency objective in addition to the standard flow matching loss. We use the same forward operator used in our other experiments to compute the consistency loss.  Further, similarly to \citet{li2024controlnet++}, the loss is only applied on timesteps $t > t_{\mathrm{min}}$ such that the consistency loss is not computed on low-quality point estimates during the initial denoising steps. While rectified flow models produce stable estimates relatively early in the denoising chain, we find in \cref{appendix:l_align_tmin_ablation} that higher $t_{\mathrm{min}}$ leads to better results, similar to the findings in \citet{li2024controlnet++}.

\textit{IT Guidance.} Inference-time guidance methods directly update the evolving sample using gradient descent on the measurement-space distance function $\mathcal{L} = \text{MSE}(\mathcal{H}(\hat{\mathbf{x}}_1),  \mathbf{y})$. We follow the FlowChef approach~\cite{patel2025flowchef} to avoid differentiating through the denoising model by computing the gradient w.r.t. $\hat{\mathbf{x}}_1$ instead of $\mathbf{x}_t$.

\subsection{3D Mesh Texturing}

\paragraph{Data Preparation.}
We follow the original TRELLIS-2 data preparation pipeline~\cite{xiang2025native} with two notable modifications. First, conditioning images are rendered using TRELLIS-2's differentiable PBR renderer rather than Blender. This ensures operator consistency: the ground-truth observation $\mathbf{y}$ in the dataset is produced by the exact same forward operator $\mathcal{H}$ that computes the feedback term during training and inference. Second, we use a fixed lighting for all assets to aid convergence in our relatively small-data regime.

\paragraph{Architecture.}
We adapt the TRELLIS-2 texture flow model, a sparse DiT operating on 32-channel PBR texture latents, by attaching LoRA adapters (rank 128, $\alpha = 128$, dropout $0.05$) to all linear layers. To inject the first-order gradient feedback $\mathbf{s}_t$, we expand the model's input projection from $2d$ to $3d$ channels, where $d = 32$ is the latent dimension, zero-initializing the additional block so the network initially behaves identically to the unguided baseline. In the look-ahead pass the gradient slot is set to $\mathbf{0}$; the refinement pass receives $\mathbf{s}_t$.

\paragraph{Feedback Signal.}
We use the first-order variant with gradients $\mathbf{\tilde{s}}_t$ taken with respect to either $\mathbf{x}_t$ or the clean estimate $\hat{\mathbf{x}}_1$.
The forward operator $\mathcal{H}$ is the composition of the frozen TRELLIS-2 PBR texture decoder and the differentiable split-sum PBR renderer (nvdiffrec~\cite{munkberg2022extracting}), and $\mathcal{L}$ is the sum-reduced MSE between the rendered output and the conditioning image. The signal is normalized per-sample by its standard deviation,
$$
\mathbf{s}_t = \mathbf{\tilde{s}}_t \;/\; \operatorname{std}(\mathbf{\tilde{s}}_t),
$$

\paragraph{Training.}
We fine-tune the LoRA layers and the expanded input layer on 7{,}500 Objaverse assets, filtered by aesthetic score $\geq 4.5$ and a maximum token count of 8{,}192 sparse voxels. Training runs for 25{,}000 steps with a batch size of 16. We use AdamW~\cite{loshchilovDecoupledWeightDecay2019} with learning rate $10^{-4}$, weight decay $0.01$, and $\beta = (0.9,\,0.95)$. Training uses bfloat16 automatic mixed precision with an EMA rate of $0.9999$. Gradients are clipped adaptively at the $95^{\text{th}}$ percentile with maximum norm $1.0$. The null-feedback probability is $p_{\text{un}} = 0.1$. Image conditioning uses a DINOv3-ViT-L/16 feature extractor~\cite{simeoni2025dinov3} at resolution 512. Training takes approximately 34 hours on 4 NVIDIA RTX PRO 6000 GPUs.

\paragraph{Inference.}
We use the default TRELLIS-2 sampler: 12 Euler steps with full two-pass execution at every step ($t_{\text{thresh}} = 1$). No CFG is applied in the main text comparisons (guidance scale 1.0), consistent with all other evaluated methods.

\paragraph{Baselines.}
\textit{Standard FT} uses identical LoRA architecture (rank 128, $\alpha=128$, dropout $0.05$) and the same training configuration as our method, using the standard flow matching objective. CFG dropout (for results shown in Tables \ref{tab:3d_results_objaverse_extended}, \ref{tab:3d_results_toys4k_extended}) is set to 0.1.

\textit{FT + $\mathcal{L}_{\text{align}}$} extends Standard FT with an additional render consistency loss, following the ControlNet++~\cite{li2024controlnet++} protocol. The loss weight is $\lambda = 5\times10^{-3}$  and the loss is applied only for $t > 0.3$, where the clean-signal estimate is sufficiently reliable; all other hyperparameters are shared with Standard FT.

\textit{IT Guidance} (FlowChef~\cite{patel2025flowchef}) is applied directly to the pretrained TRELLIS-2 model without any fine-tuning. At each of the 12 ODE steps, one SGD update with learning rate ($s'$) $0.1$ is applied to the noisy latent $\mathbf{x}_t$ using the MSE render loss gradient, following the official implementation defaults.

Results for baselines with additional configurations, including number of sampling steps, CFG, consistency loss weight $\lambda$, and guidance learning rate $s'$ can be seen in Tables \ref{tab:3d_results_objaverse_extended}, \ref{tab:3d_results_toys4k_extended}.

\section{Additional Results}
\label{appendix:additional_results}

\subsection{Image-to-Image Translation}
\label{appendix:image_to_image}
We provide additional qualitative comparisons for the JPEG restoration task in \cref{appendix:fig:jpeg_qualitative}, depth-to-RGB in \cref{appendix:fig:depth_qualitative}, and super resolution in \cref{appendix:fig:sr_qualitative}. Tables with full results and additional settings, such as number of sampling steps, Classifier-Free Guidance, and hyperparameter choices in \cref{appendix:tab:depth_to_rgb}, \cref{appendix:tab:edge_to_rgb}, \cref{appendix:tab:jpeg_restore}, and \cref{appendix:tab:super_resolution}.

\paragraph{Extended Experimental Analysis.} 
While the main text evaluates the configurations marked with $(*)$, this section provides a comprehensive comparison across varying sampling budgets and guidance strengths. 

For open-loop baselines, standard test-time enhancement techniques fail to resolve the fundamental fidelity and plausibility shortcomings. Doubling the sampling budget ($2\times$ steps) yields negligible improvements in fidelity metrics. Similarly, applying CFG to these baselines often degrades plausibility (FID) without significantly bridging the fidelity gap. These results confirm that open-loop failures cannot be resolved with standard test-time enhancements. 

In contrast, our proposed CFG scheme (Sec.~\ref{sec:inference}) often boosts \algname's fidelity. For example, as shown in Table~\ref{appendix:tab:super_resolution}, $w=3.0$ guidance improves our zero-order variant's PSNR by 6.0~dB, compared to a negligible 0.3~dB for the baseline.

Inference-time guidance (\textit{IT Guidance}) shows a strict trade-off between plausibility and fidelity across all tasks depending on hyperparameter choices while our method consistently achieves the best of both worlds.

\begin{figure}[t]
    \centering
    \includegraphics[width=\linewidth]{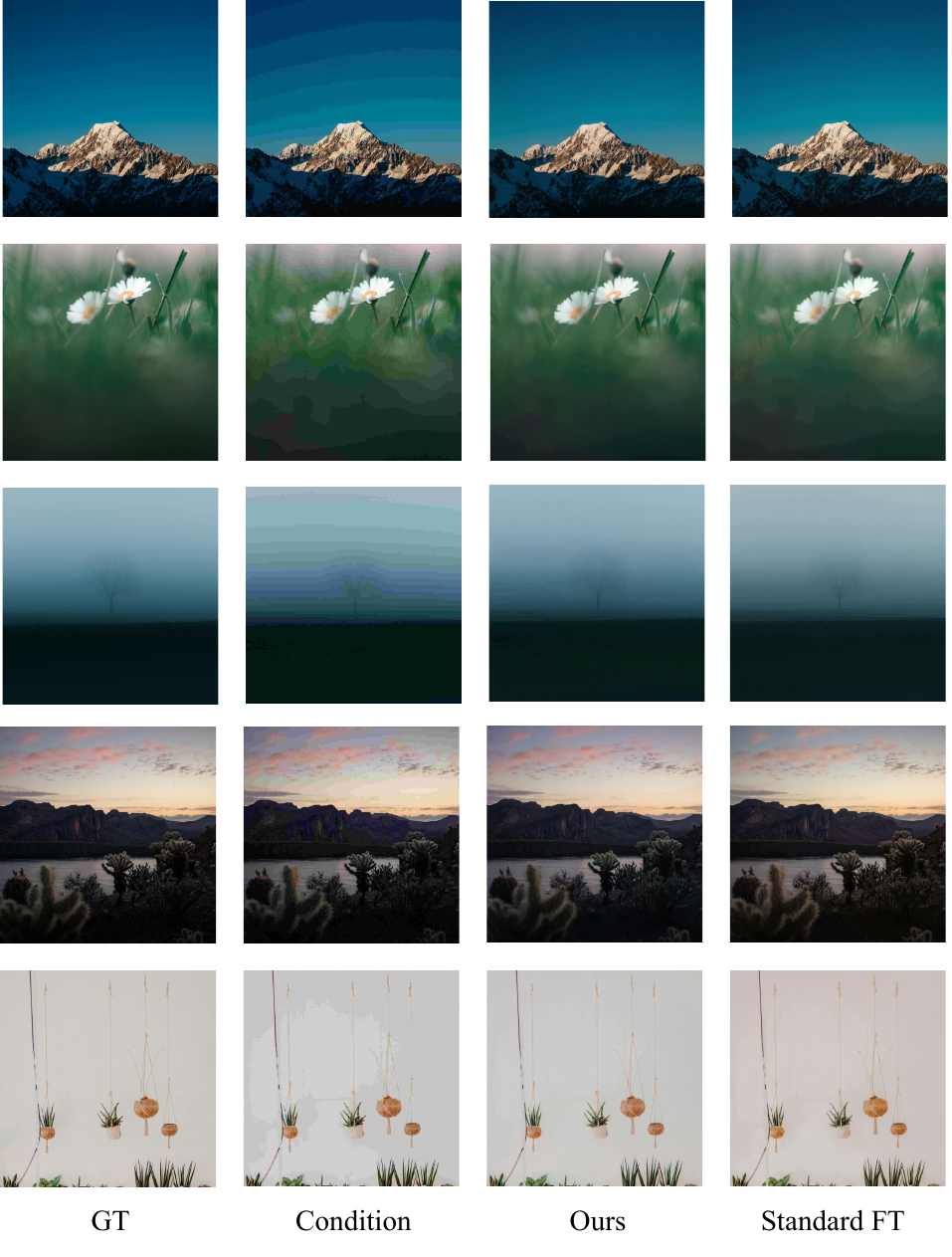}
    \caption{\textbf{Qualitative comparison for the JPEG restoration task.} Our method reduces color banding quantization artifacts (rows 1-3) and color shifts (rows 4-5).}
    \label{appendix:fig:jpeg_qualitative}
\end{figure}

\begin{figure}[t]
    \centering
    \includegraphics[width=\linewidth]{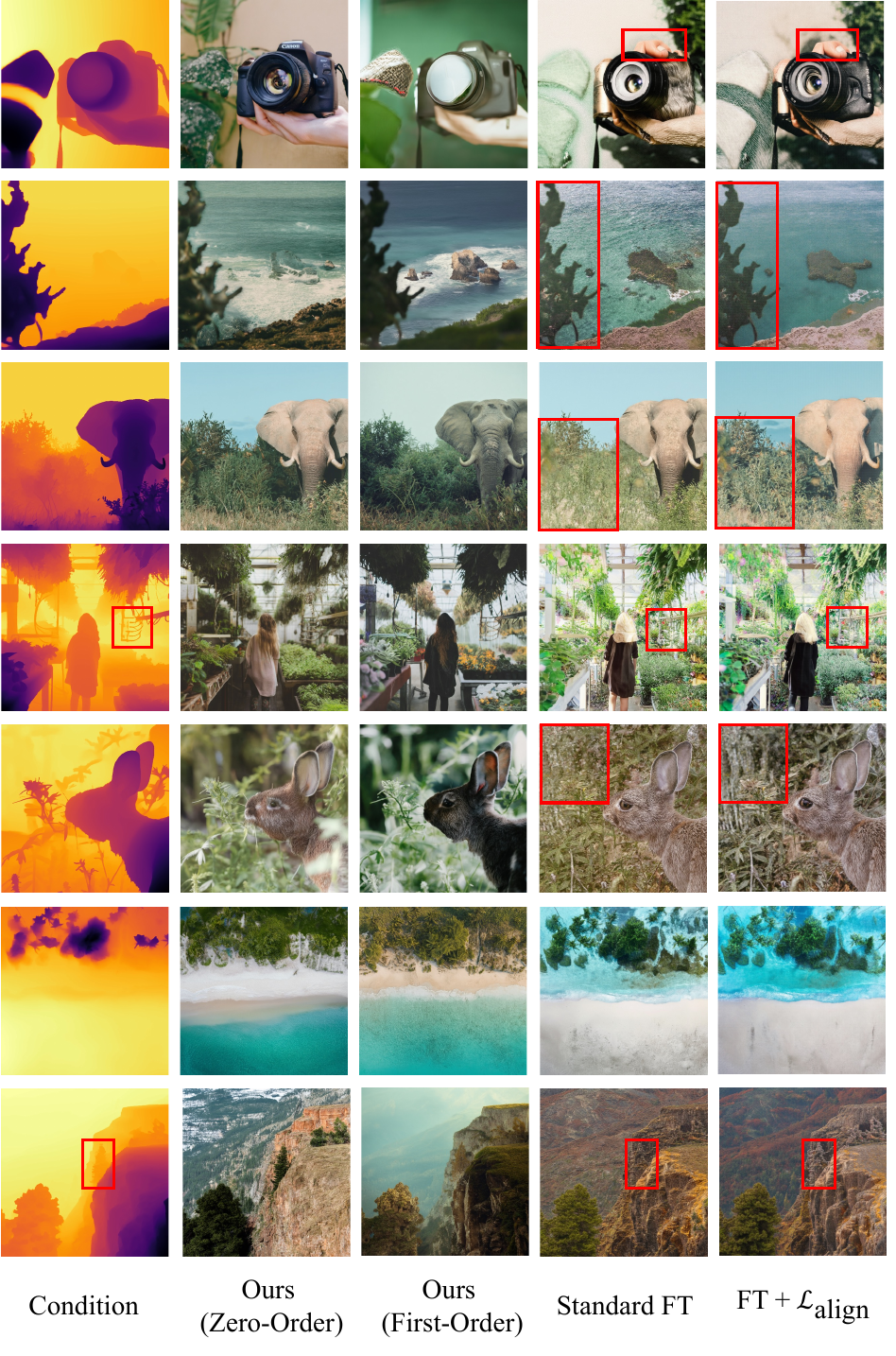}
    \caption{Qualitative comparison for the depth-to-RGB task.}
    \label{appendix:fig:depth_qualitative}
\end{figure}

\begin{figure}[t]
    \centering
    \includegraphics[width=\linewidth]{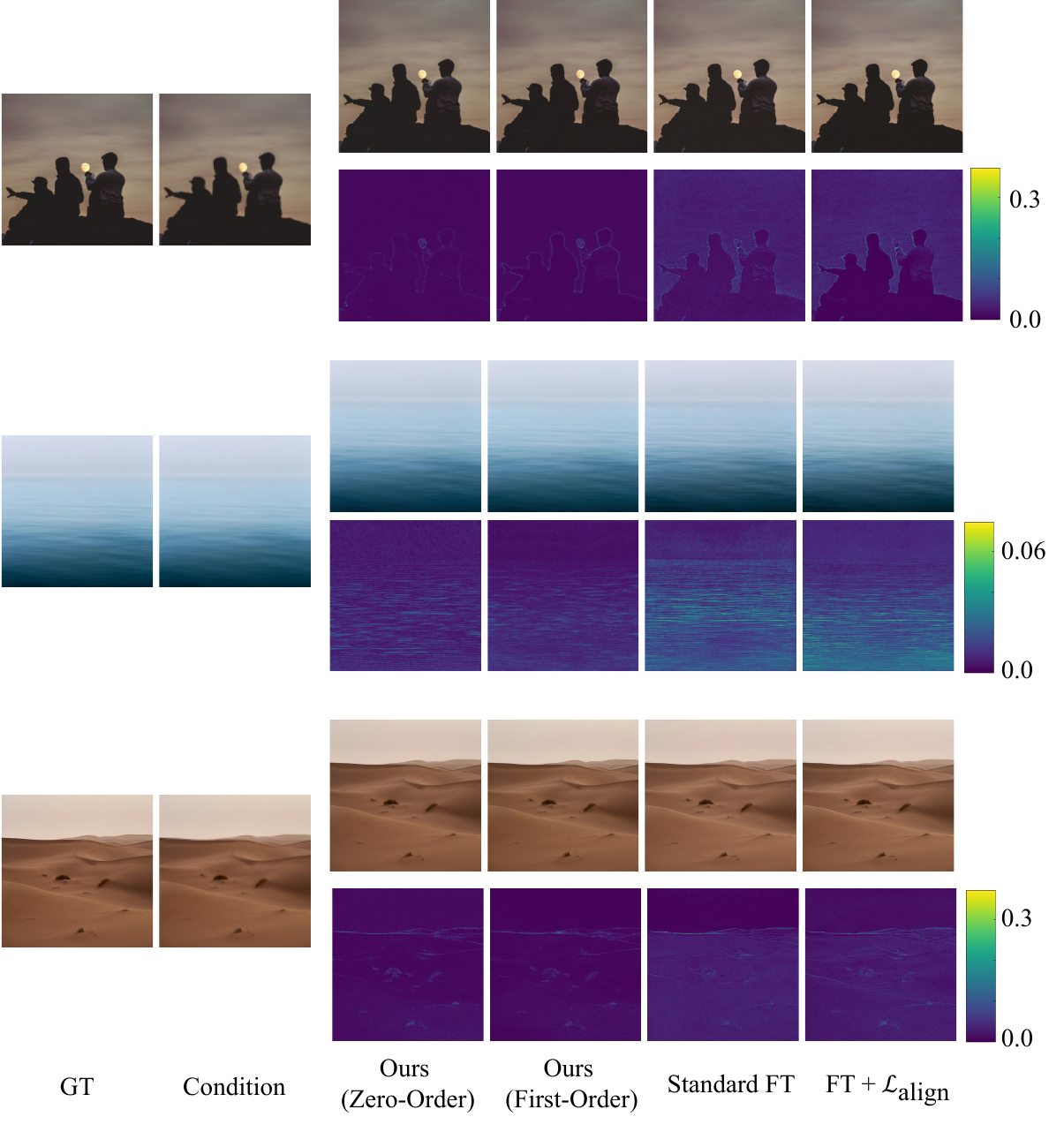}
    \caption{\textbf{Qualitative comparison for the super-resolution task.} Error maps show per-pixel MAE.}
    \label{appendix:fig:sr_qualitative}
\end{figure}

\begin{table}[t]
\centering
\caption{Ablation of $t_\mathrm{min}$ parameter in $\mathcal{L}_{\text{align}}$ baseline on depth task.}
\label{appendix:l_align_tmin_ablation}
\adjustbox{max width=\linewidth}{%
\begin{tabular}{lccc}
  \toprule
   & \multicolumn{2}{c}{Fidelity} & \multicolumn{1}{c}{Plausibility} \\
  \cmidrule(lr){2-3} \cmidrule(lr){4-4}
  $t_\mathrm{min}$ & MAE $\downarrow$ & $\delta_1$ $\uparrow$ & FID $\downarrow$ \\
  \midrule
  0.8 & 0.0834 & 0.7933 & 17.57 \\
  0.5 & 0.0799 & 0.7854 & 21.31 \\
  \bottomrule
\end{tabular}
}
\end{table}

\begin{table}[t]
\centering
\caption{Super resolution results. Asterisks (*) denote base configurations compared in the main text. Shaded rows indicate \algname variants.}
\label{appendix:tab:super_resolution}
\adjustbox{max width=\linewidth}{%
\begin{tabular}{lcccc}
  \toprule
   & \multicolumn{3}{c}{Fidelity} & \multicolumn{1}{c}{Plausibility} \\
  \cmidrule(lr){2-4} \cmidrule(lr){5-5}
  Method & PSNR $\uparrow$ & SSIM $\uparrow$ & LPIPS $\downarrow$ & FID $\downarrow$ \\
  \midrule
  Standard FT (*) & 34.35$_{\scriptstyle\pm 2.78}$ & 96.88$_{\scriptstyle\pm 4.32}$ & 0.83$_{\scriptstyle\pm 0.43}$ & 3.93 \\
  \quad + CFG ($w=3.0$) & 34.65$_{\scriptstyle\pm 2.94}$ & 97.23$_{\scriptstyle\pm 3.29}$ & 0.91$_{\scriptstyle\pm 0.49}$ & 4.09 \\
  \quad + CFG ($w=5.0$) & 34.07$_{\scriptstyle\pm 3.02}$ & 97.17$_{\scriptstyle\pm 2.76}$ & 1.15$_{\scriptstyle\pm 0.68}$ & 4.57 \\
  \quad + $2\times$ steps & 34.46$_{\scriptstyle\pm 2.74}$ & 96.93$_{\scriptstyle\pm 3.84}$ & 0.86$_{\scriptstyle\pm 0.44}$ & 4.07 \\
  \midrule
  Standard FT + $\mathcal{L}_{\text{align}}$ (*) & 35.21$_{\scriptstyle\pm 2.80}$ & 97.53$_{\scriptstyle\pm 2.96}$ & 0.79$_{\scriptstyle\pm 0.36}$ & 4.11 \\
  \quad + $2\times$ steps & 35.11$_{\scriptstyle\pm 2.81}$ & 97.43$_{\scriptstyle\pm 2.74}$ & 0.83$_{\scriptstyle\pm 0.39}$ & 4.10 \\
  \midrule
  IT Guidance ($\lambda = 0.1$) & 38.30$_{\scriptstyle\pm 2.80}$ & 97.20$_{\scriptstyle\pm 1.91}$ & 1.33$_{\scriptstyle\pm 1.46}$ & 20.93 \\
  IT Guidance ($\lambda = 0.5$) (*) & 43.02$_{\scriptstyle\pm 4.85}$ & 98.29$_{\scriptstyle\pm 3.19}$ & 0.65$_{\scriptstyle\pm 2.24}$ & 18.96 \\
  IT Guidance ($\lambda = 0.5$, $2\times$ steps) & \textbf{46.05}$_{\scriptstyle\pm 4.91}$ & 98.95$_{\scriptstyle\pm 2.72}$ & 0.27$_{\scriptstyle\pm 1.18}$ & 18.87 \\
  \midrule
  \rowcolor{gray!10} Zero-order (*) & 39.25$_{\scriptstyle\pm 2.46}$ & 98.18$_{\scriptstyle\pm 4.06}$ & 0.21$_{\scriptstyle\pm 0.11}$ & \textbf{3.36} \\
  \rowcolor{gray!10} \quad + CFG ($w=3.0$) & 45.25$_{\scriptstyle\pm 1.35}$ & \textbf{99.08}$_{\scriptstyle\pm 1.68}$ & \textbf{0.13}$_{\scriptstyle\pm 0.15}$ & 3.58 \\
  \rowcolor{gray!10} \quad + CFG ($w=5.0$) & 42.61$_{\scriptstyle\pm 1.61}$ & 98.70$_{\scriptstyle\pm 1.10}$ & 0.34$_{\scriptstyle\pm 0.38}$ & 4.20 \\
  \midrule
  \rowcolor{gray!10} First-order (w.r.t. $\mathbf{x}_t$) (*) & 36.27$_{\scriptstyle\pm 3.95}$ & 97.10$_{\scriptstyle\pm 4.69}$ & 0.66$_{\scriptstyle\pm 0.39}$ & 4.30 \\
  \rowcolor{gray!10} First-order (w.r.t. $\hat{\mathbf{x}}_1$) (*) & 39.21$_{\scriptstyle\pm 3.85}$ & 97.64$_{\scriptstyle\pm 5.01}$ & 0.43$_{\scriptstyle\pm 0.30}$ & 3.83 \\
  \rowcolor{gray!10} \quad + CFG ($w=3.0$) & 45.14$_{\scriptstyle\pm 2.98}$ & 98.95$_{\scriptstyle\pm 2.50}$ & 0.17$_{\scriptstyle\pm 0.17}$ & 4.74 \\
  \rowcolor{gray!10} \quad + CFG ($w=5.0$) & 44.07$_{\scriptstyle\pm 2.11}$ & 98.96$_{\scriptstyle\pm 2.04}$ & 0.33$_{\scriptstyle\pm 0.55}$ & 6.45 \\
  \midrule
  \rowcolor{gray!10} Combined (w.r.t. $\mathbf{x}_t$) (*) & 39.95$_{\scriptstyle\pm 2.77}$ & 98.25$_{\scriptstyle\pm 4.18}$ & 0.21$_{\scriptstyle\pm 0.12}$ & 3.40 \\
  \rowcolor{gray!10} \quad + CFG ($w=3.0$) & 43.77$_{\scriptstyle\pm 1.40}$ & 98.79$_{\scriptstyle\pm 2.58}$ & 0.15$_{\scriptstyle\pm 0.17}$ & 3.62 \\
  \rowcolor{gray!10} \quad + CFG ($w=5.0$) & 41.43$_{\scriptstyle\pm 1.57}$ & 98.25$_{\scriptstyle\pm 2.10}$ & 0.42$_{\scriptstyle\pm 0.49}$ & 4.39 \\
  \rowcolor{gray!10} Combined (w.r.t. $\hat{\mathbf{x}}_1$) (*) & 39.77$_{\scriptstyle\pm 2.83}$ & 98.24$_{\scriptstyle\pm 3.94}$ & 0.23$_{\scriptstyle\pm 0.13}$ & 3.39 \\
  \bottomrule
\end{tabular}
}
\end{table}

\begin{table}[t]
\centering
\caption{\textbf{Depth-to-RGB generation results.} Asterisks (*) denote base configurations compared in the main text. Shaded rows indicate \algname variants.}
\label{appendix:tab:depth_to_rgb}
\adjustbox{max width=\linewidth}{%
\begin{tabular}{lccc}
  \toprule
   & \multicolumn{2}{c}{Fidelity} & \multicolumn{1}{c}{Plausibility} \\
  \cmidrule(lr){2-3} \cmidrule(lr){4-4}
  Method & MAE $\downarrow$ & $\delta_1$ $\uparrow$ & FID $\downarrow$ \\
  \midrule
  Standard FT (*)                         & 0.0847$_{\scriptstyle\pm 0.0593}$ & 0.7886$_{\scriptstyle\pm 0.1930}$ & 18.21           \\
  \quad + CFG ($w=3.0$)              & 0.0814$_{\scriptstyle\pm 0.0394}$ & 0.8003$_{\scriptstyle\pm 0.1331}$ & 14.90           \\
  \quad + CFG ($w=5.0$)              & 0.0823$_{\scriptstyle\pm 0.0388}$ & 0.7959$_{\scriptstyle\pm 0.1333}$ & 15.77           \\
  \quad + $2\times$ steps            & 0.0844$_{\scriptstyle\pm 0.0415}$ & 0.7907$_{\scriptstyle\pm 0.1381}$ & 16.74           \\
 FT + $\mathcal{L}_{\text{align}}$ (*)    & 0.0837$_{\scriptstyle\pm 0.0600}$ & 0.7930$_{\scriptstyle\pm 0.1926}$ & 17.57           \\
  \midrule
  IT Guidance ($\lambda=0.0001$) & 0.1617$_{\scriptstyle\pm 0.0787}$ & 0.5919$_{\scriptstyle\pm 0.1986}$ & 23.42 \\
  IT Guidance ($\lambda=0.001$) (*) & 0.0866$_{\scriptstyle\pm 0.0360}$ & 0.7592$_{\scriptstyle\pm 0.1284}$ & 223.54 \\
  IT Guidance ($\lambda=0.01$) & 0.1385$_{\scriptstyle\pm 0.0501}$ & 0.6145$_{\scriptstyle\pm 0.1378}$ & 328.57 \\
  \midrule
  \rowcolor{gray!10} Zero-order (*)                        & 0.0763$_{\scriptstyle\pm 0.0563}$ & 0.8188$_{\scriptstyle\pm 0.1843}$ & 15.70           \\
  \rowcolor{gray!10} First-order (w.r.t. $\mathbf{x}_t$) (*)         & 0.0747$_{\scriptstyle\pm 0.0535}$ & 0.8268$_{\scriptstyle\pm 0.1726}$ & \textbf{14.57}  \\
  \rowcolor{gray!10} \quad + CFG ($w=3.0$)              & 0.0616$_{\scriptstyle\pm 0.0465}$ & 0.8705$_{\scriptstyle\pm 0.1474}$ & 14.62           \\
  \rowcolor{gray!10} \quad + CFG ($w=5.0$)              & \textbf{0.0534}$_{\scriptstyle\pm 0.0422}$ & \textbf{0.8961}$_{\scriptstyle\pm 0.1280}$ & 14.76           \\
  \rowcolor{gray!10} First-order (w.r.t. $\hat{\mathbf{x}}_1$) (*)   & 0.0818$_{\scriptstyle\pm 0.0582}$ & 0.7973$_{\scriptstyle\pm 0.1933}$ & 15.89           \\
  \rowcolor{gray!10} Combined (w.r.t. $\hat{\mathbf{x}}_1$) (*)      & 0.0829$_{\scriptstyle\pm 0.0610}$ & 0.7949$_{\scriptstyle\pm 0.1953}$ & 15.93           \\
  \rowcolor{gray!10} Combined (w.r.t. $\mathbf{x}_t$) (*)            & 0.0783$_{\scriptstyle\pm 0.0606}$ & 0.8175$_{\scriptstyle\pm 0.1878}$ & 15.39           \\
  \bottomrule
\end{tabular}
}
\end{table}

\begin{table}[t]
\centering
\caption{\textbf{JPEG restoration results.} Asterisks (*) denote base configurations compared in the main text. Shaded rows indicate \algname variants.}
\label{appendix:tab:jpeg_restore}
\adjustbox{max width=\linewidth}{%
\begin{tabular}{lcccc}
  \toprule
   & \multicolumn{3}{c}{Fidelity} & \multicolumn{1}{c}{Plausibility} \\
  \cmidrule(lr){2-4} \cmidrule(lr){5-5}
  Method & PSNR & SSIM & LPIPS & FID \\
  \midrule
  Standard FT (*) & 26.29$_{\scriptstyle\pm 2.91}$ & 79.45$_{\scriptstyle\pm 10.59}$ & 22.24$_{\scriptstyle\pm 5.44}$ & 4.35 \\
  \quad + CFG. ($w=3.0$) & 26.56$_{\scriptstyle\pm 3.04}$ & 80.32$_{\scriptstyle\pm 10.18}$ & 21.62$_{\scriptstyle\pm 5.48}$ &  4.36 \\
  \quad + CFG. ($w=5.0$) & 26.58$_{\scriptstyle\pm 3.10}$ & 80.85$_{\scriptstyle\pm 9.89}$ & 21.44$_{\scriptstyle\pm 5.47}$ &  4.62 \\
  \quad + $2\times$ steps & 26.30$_{\scriptstyle\pm 3.00}$ & 79.17$_{\scriptstyle\pm 10.54}$ & 22.59$_{\scriptstyle\pm 5.58}$  & 4.30 \\
  \midrule
  \rowcolor{gray!10} Zero-order (*) & 28.86$_{\scriptstyle\pm 4.00}$ & 83.13$_{\scriptstyle\pm 9.64}$ & 16.33$_{\scriptstyle\pm 4.59}$ &  \textbf{3.80} \\
  \rowcolor{gray!10} \quad + CFG. ($w=3.0$) & \textbf{29.79}$_{\scriptstyle\pm 4.21}$ & 85.58$_{\scriptstyle\pm 8.50}$ & \textbf{14.72}$_{\scriptstyle\pm 4.93}$  & 3.85 \\
  \rowcolor{gray!10} \quad + CFG. ($w=5.0$) & 29.66$_{\scriptstyle\pm 4.07}$ & \textbf{85.83}$_{\scriptstyle\pm 8.02}$ & 16.19$_{\scriptstyle\pm 5.95}$ & 4.58 \\
  \bottomrule
\end{tabular}
}
\end{table}

\begin{table}[t]
\centering
\caption{\textbf{Edge-to-RGB generation results.} Asterisks (*) denote base configurations compared in the main text. Shaded rows indicate \algname variants.}
\label{appendix:tab:edge_to_rgb}
\adjustbox{max width=\linewidth}{%
\begin{tabular}{lcccccc}
  \toprule
   & \multicolumn{2}{c}{Fidelity} & \multicolumn{1}{c}{Plausibility} \\
  \cmidrule(lr){2-3} \cmidrule(lr){4-4}
  Method & Edge MAE $\downarrow$ & Edge MSE $\downarrow$ & FID $\downarrow$ \\
  \midrule
  Standard FT (*) &0.0533$_{\scriptstyle\pm 0.0286}$ & 0.0137$_{\scriptstyle\pm 0.0087}$ & 13.98 \\
  \quad + CFG ($w=3.0$) & 0.0500$_{\scriptstyle\pm 0.0286}$ & 0.0137$_{\scriptstyle\pm 0.0092}$ & \textbf{12.41} \\ 
  \quad + CFG ($w=5.0$) & 0.0513$_{\scriptstyle\pm 0.0296}$ & 0.0150$_{\scriptstyle\pm 0.0101}$ & 13.64 \\
  \quad  $2\times$ steps &0.0512$_{\scriptstyle\pm 0.0285}$ & 0.0134$_{\scriptstyle\pm 0.0087}$ & 13.17 \\
  \midrule
  Standard FT + $\mathcal{L}_{\text{align}}$ (*) & 0.0501$_{\scriptstyle\pm 0.0277}$ & 0.0128$_{\scriptstyle\pm 0.0087}$ & 14.47 \\
  \midrule
  IT Guidance ($\lambda = 0.001$) & 0.0620$_{\scriptstyle\pm 0.0359}$ & 0.0247$_{\scriptstyle\pm 0.0167}$ & 23.97 \\
  IT Guidance ($\lambda = 0.01$) (*) & 0.0416$_{\scriptstyle\pm 0.0235}$ & 0.0129$_{\scriptstyle\pm 0.0093}$ & 97.65 \\
  \midrule
  \rowcolor{gray!10} Zero-order (*) & 0.0456$_{\scriptstyle\pm 0.0253}$ & 0.0104$_{\scriptstyle\pm 0.0072}$ & 14.81 \\
  \rowcolor{gray!10} First-order (w.r.t. $\mathbf{x}_t$) (*) & 0.0460$_{\scriptstyle\pm 0.0265}$ & 0.0111$_{\scriptstyle\pm 0.0077}$ & 14.29 \\
  \rowcolor{gray!10} First-order (w.r.t. $\hat{\mathbf{x}}_1$) (*) & 0.0435$_{\scriptstyle\pm 0.0294}$ & 0.0123$_{\scriptstyle\pm 0.0089}$ & 14.97 \\
  \rowcolor{gray!10} Combined (w.r.t. $\mathbf{x}_t$) (*) & 0.0435$_{\scriptstyle\pm 0.0253}$ & 0.0097$_{\scriptstyle\pm 0.0072}$ & 13.91 \\
  \rowcolor{gray!10} Combined (w.r.t. $\hat{\mathbf{x}}_1$) (*)  & 0.0408$_{\scriptstyle\pm 0.0232}$ & 0.0085$_{\scriptstyle\pm 0.0060}$ & 13.68 \\
  \rowcolor{gray!10}   \quad + CFG ($w=3.0$) & 0.0322$_{\scriptstyle\pm 0.0201}$ & \textbf{0.0057}$_{\scriptstyle\pm 0.0053}$ & 14.62 \\
  \rowcolor{gray!10}   \quad + CFG ($w=5.0$) & \textbf{0.0320}$_{\scriptstyle\pm 0.0220}$ & 0.0062$_{\scriptstyle\pm 0.0071}$ & 17.97 \\
  \bottomrule
\end{tabular}
}
\end{table}

\paragraph{FlowChef limitations.}
While \citet{patel2025flowchef} show that their inference-time guidance scheme works for simple forward operators such as inpainting and super-resolution, we find that FlowChef does not generalize to complex forward operators such as neural networks. To validate this, we performed a dense sweep over key hyperparameters, including learning rate $\lambda \in \{1\times10^{-5},\,2.5\times10^{-5},\,5\times10^{-5},\,7.5\times10^{-5},\,1\times10^{-4},\,2.5\times10^{-4},\,5\times10^{-4},\,1\times10^{-3},\,5\times10^{-3},\,1\times10^{-2},\,2\times10^{-2},\,5\times10^{-2},\,1\times10^{-1},\,5\times10^{-1},\,1.0\}$, the percentage of steps applying guidance ($80\%, 100\%$), and total number of steps ($40, 80, 100$).
Despite these extensive attempts, we could not find any setting that resulted in both satisfactory visual quality and adherence to the conditioning for the Edge and Depth tasks. We provide qualitative examples of these failures in \cref{appendix:fig:flowchef_neuralnetwork_failures}.

\begin{figure}[t]
    \centering
    \begin{subfigure}[b]{0.48\linewidth}
        \centering
        \includegraphics[width=\linewidth]{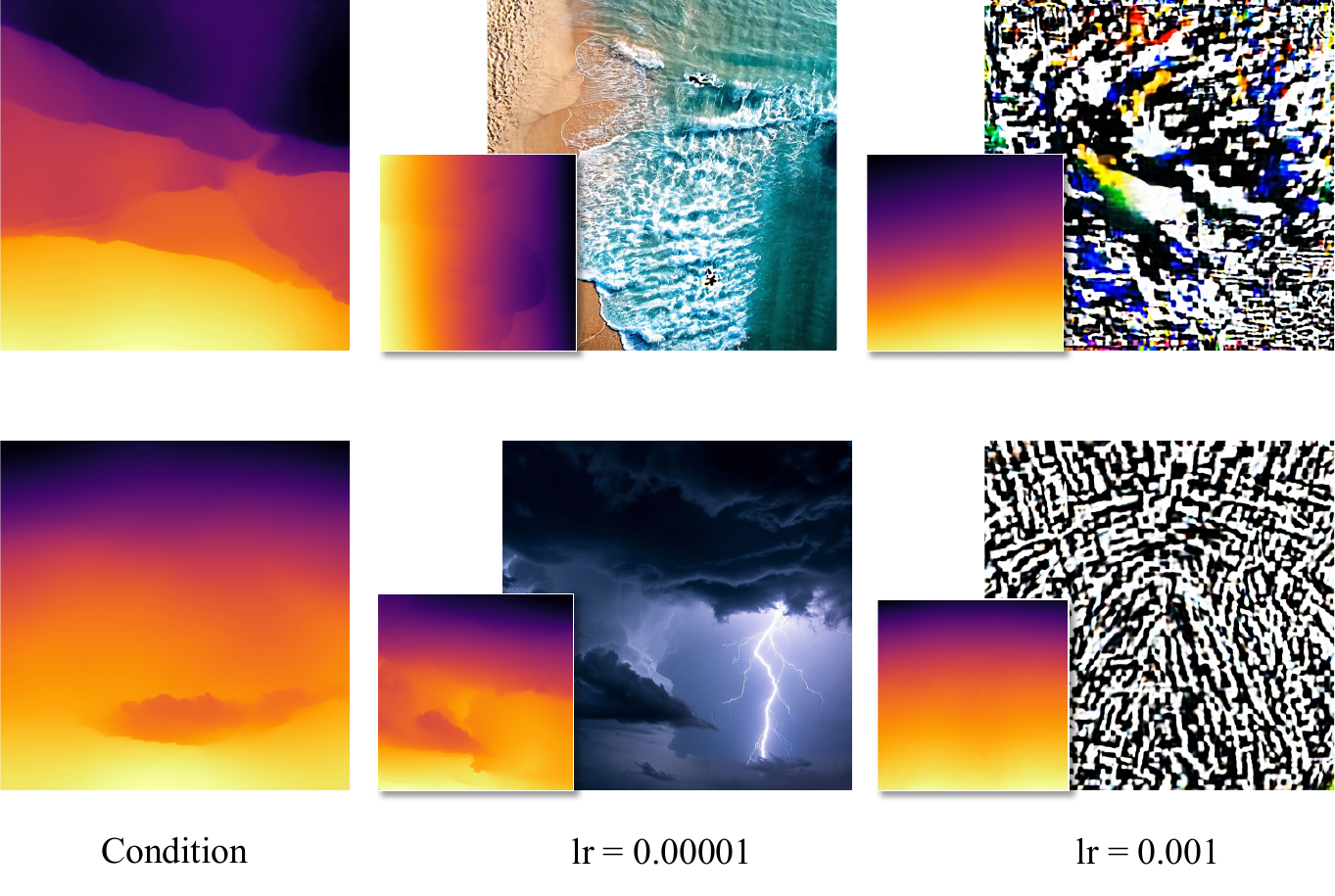}
        \caption{Depth}
    \end{subfigure}
    \hfill
    \begin{subfigure}[b]{0.48\linewidth}
        \centering
        \includegraphics[width=\linewidth]{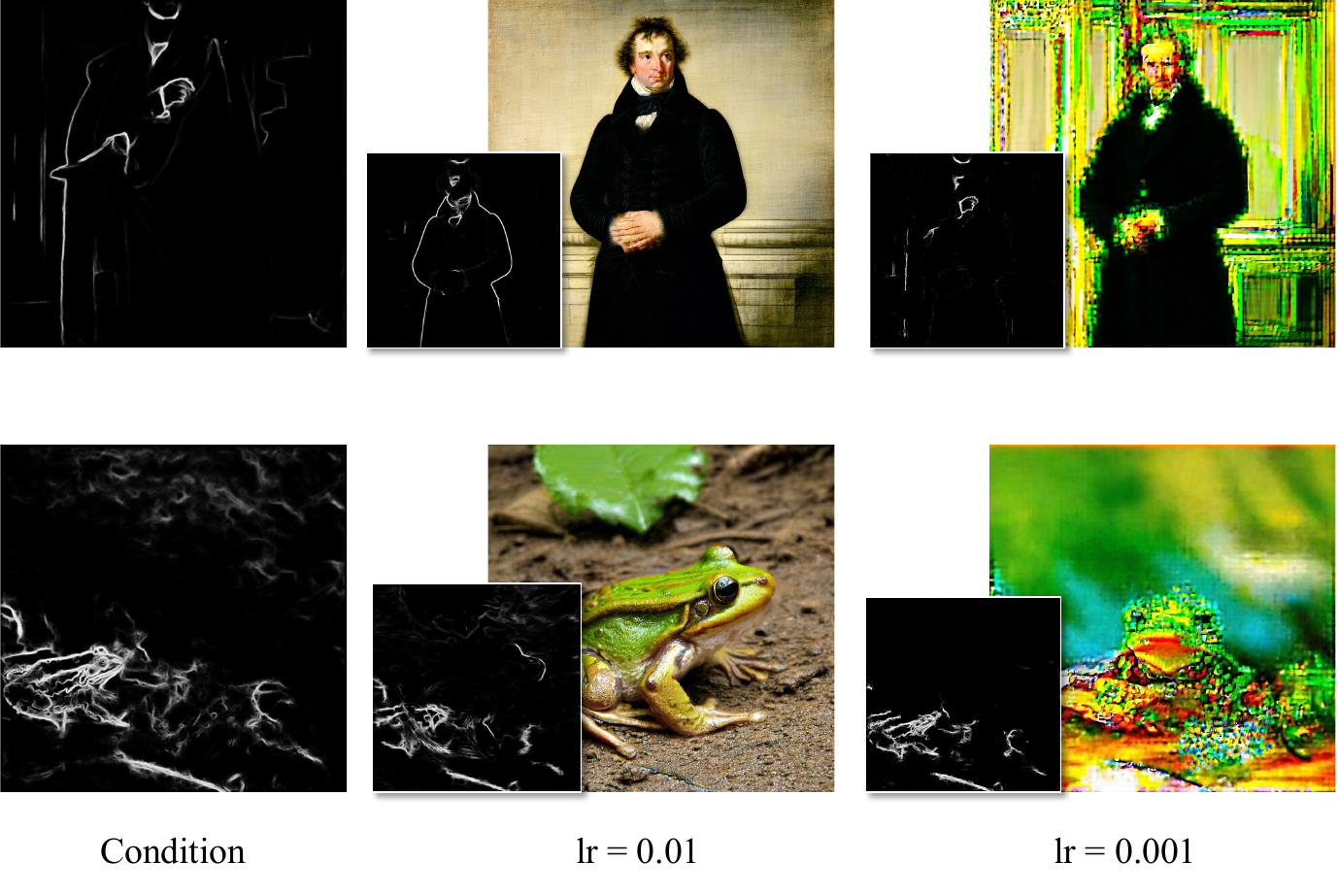}
        \caption{Edge}
    \end{subfigure}
    \caption{\textbf{Qualitative examples of failures of FlowChef for neural-network forward operators.} As the adherence to the condition improves, shown as insets, the visual quality degrades significantly.}
    \label{appendix:fig:flowchef_neuralnetwork_failures}
\end{figure}

\subsection{3D Mesh Texturing}

\begin{figure}
    \centering
    \includegraphics[width=1\linewidth]{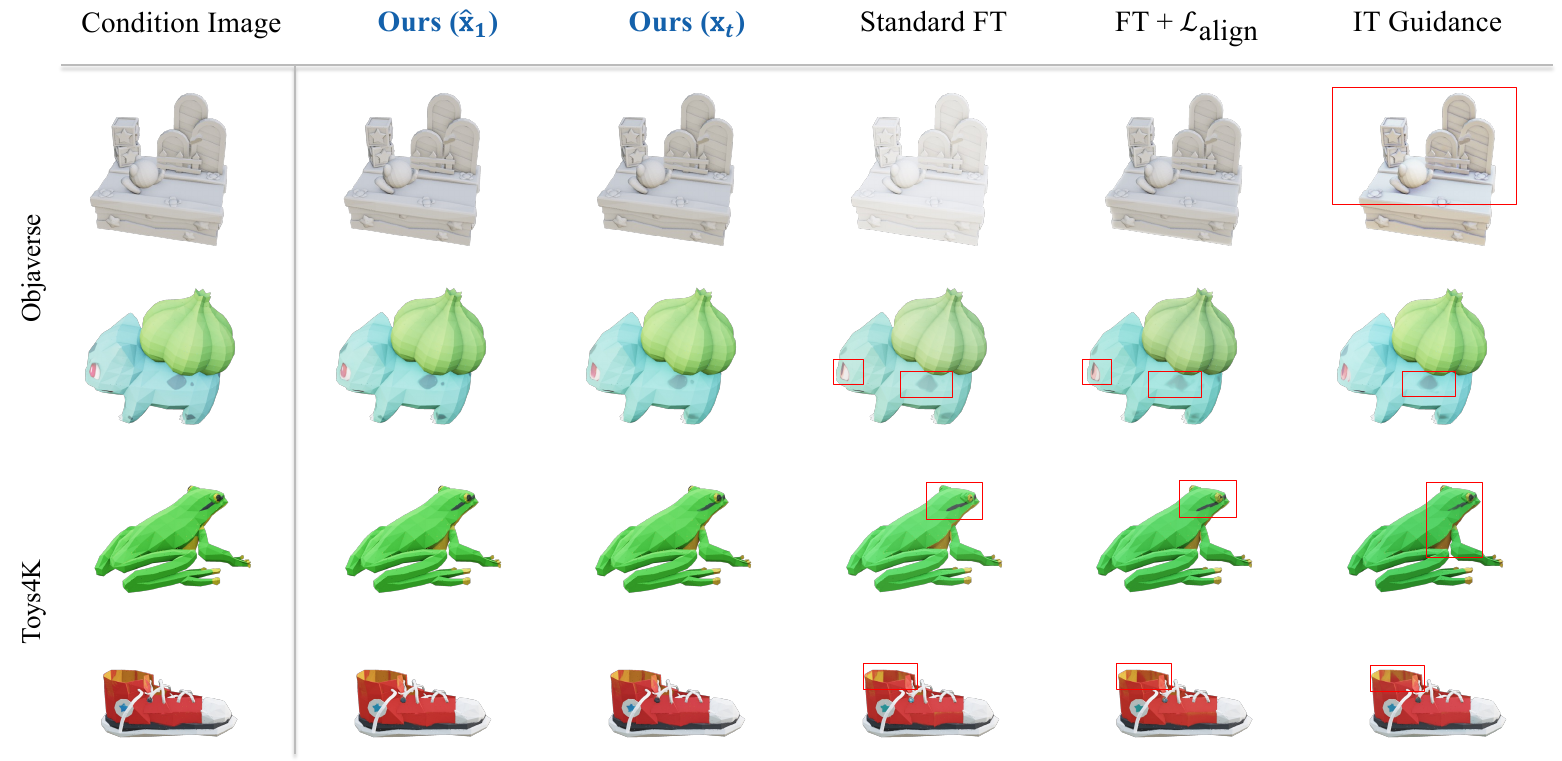}
    \caption{3D Mesh Texturing Results (Objaverse, Toys4K).}
    \label{fig:texturing_additional}
\end{figure}

\begin{figure}
    \centering
    \includegraphics[width=1\linewidth]{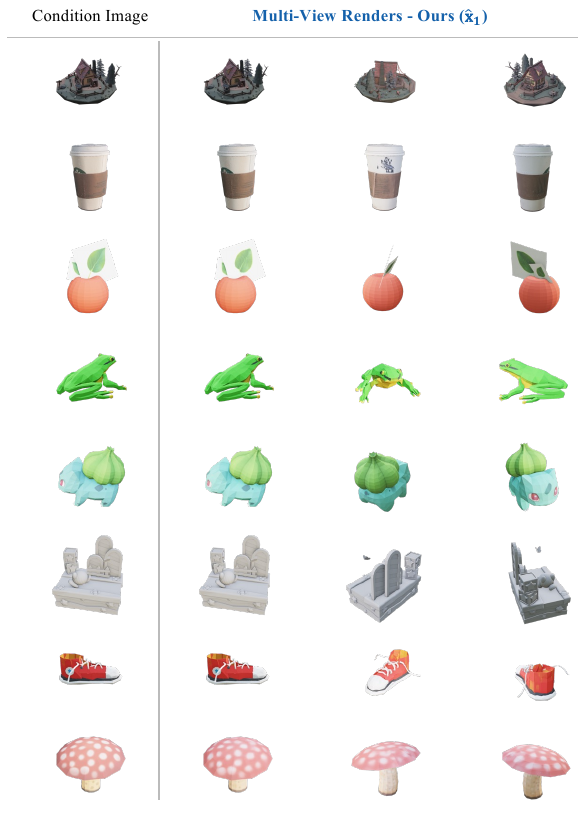}
    \caption{Multi-view visualizations of textured objects (Objaverse, Toys4K).}
    \label{fig:texturing_multi_view}
\end{figure}

We present extended results in Tables~\ref{tab:3d_results_objaverse_extended} and~\ref{tab:3d_results_toys4k_extended}, additional qualitative comparisons in Figure~\ref{fig:texturing_additional}, and multi-view visualizations in Figure~\ref{fig:texturing_multi_view}.

The patterns observed in image-to-image tasks (Appendix \ref{appendix:image_to_image}) persist in the 3D domain (Tables~\ref{tab:3d_results_objaverse_extended} and~\ref{tab:3d_results_toys4k_extended}). For open-loop baselines, standard enhancements like doubling sampling steps or applying CFG prove suboptimal or even detrimental; for example, applying CFG ($w=3.0$) to Standard FT on Toys4K leads to a 1.57~dB drop in M.PSNR and a significant plausibility collapse (FID increases from 8.87 to 10.30). While our proposed CFG scheme for \algname (Sec.~\ref{sec:inference}) provides further fidelity and plausibility gains---such as a 1.04~dB M.PSNR and 0.65 FID improvement on Toys4K---we emphasize that \algname without any explicit guidance already significantly outperforms baselines using doubled sampling budgets or CFG. This indicates that while \algname can effectively leverage optional user-controlled test-time guidance, its primary strength lies in the learned correction policy which internalizes the feedback signal.

\begin{table*}[t]
\centering
\caption{\textbf{Extended Quantitative Results for 3D Texturing (Objaverse).} Asterisks (*) denote configurations included in the main text comparisons. Shaded rows indicate \algname variants.}
\label{tab:3d_results_objaverse_extended}
\adjustbox{max width=\linewidth}{%
\begin{tabular}{l cccc ccccc}
\toprule
& \multicolumn{4}{c}{\textbf{Fidelity}} & \multicolumn{5}{c}{\textbf{Plausibility}} \\
\cmidrule(lr){2-5} \cmidrule(lr){6-10}
Method & M. PSNR $\uparrow$ & SSIM $\uparrow$ & LPIPS $\downarrow$ & CLIP $\uparrow$ & MV-M.PSNR $\uparrow$ & MV-SSIM $\uparrow$ & MV-LPIPS $\downarrow$ & MV-CLIP $\uparrow$ & FID $\downarrow$ \\
\midrule
IT Guidance (*) &  25.87{\scriptsize$\pm$3.7} & 0.986{\scriptsize$\pm$0.015} & 0.0191{\scriptsize$\pm$0.0250} & 0.973{\scriptsize$\pm$0.029} & 22.79{\scriptsize$\pm$5.7} & 0.989{\scriptsize$\pm$0.011} & 0.0124{\scriptsize$\pm$0.0113} & 0.968{\scriptsize$\pm$0.024} & 9.10 \\
\quad + $2\times$ steps &  \textbf{27.05{\scriptsize$\pm$3.5}} & \underline{0.988{\scriptsize$\pm$0.013}} & 0.0174{\scriptsize$\pm$0.0231} & 0.978{\scriptsize$\pm$0.026} & 23.47{\scriptsize$\pm$5.9} & 0.990{\scriptsize$\pm$0.011} & 0.0122{\scriptsize$\pm$0.0114} & 0.969{\scriptsize$\pm$0.026} & 8.70 \\
\quad + $s = 0.5$ &  26.77{\scriptsize$\pm$3.2} & 0.987{\scriptsize$\pm$0.015} & 0.0218{\scriptsize$\pm$0.0276} & 0.970{\scriptsize$\pm$0.035} & 23.04{\scriptsize$\pm$6.7} & 0.989{\scriptsize$\pm$0.011} & 0.0136{\scriptsize$\pm$0.0130} & 0.966{\scriptsize$\pm$0.025} & 10.11 \\
\midrule
Standard FT (*) & 21.91{\scriptsize$\pm$3.8} & 0.981{\scriptsize$\pm$0.020} & 0.0189{\scriptsize$\pm$0.0178} & 0.970{\scriptsize$\pm$0.022} & 23.05{\scriptsize$\pm$6.3} & 0.990{\scriptsize$\pm$0.012} & 0.0116{\scriptsize$\pm$0.0114} & 0.970{\scriptsize$\pm$0.022} & 8.74 \\
\quad + CFG ($w = 3$) & 20.41{\scriptsize$\pm$4.4} & 0.980{\scriptsize$\pm$0.023} & 0.0217{\scriptsize$\pm$0.0215} & 0.961{\scriptsize$\pm$0.030} & 21.06{\scriptsize$\pm$5.5} & 0.988{\scriptsize$\pm$0.015} & 0.0134{\scriptsize$\pm$0.0136} & 0.963{\scriptsize$\pm$0.024} & 10.11 \\
\quad + CFG ($w = 5$) & 19.45{\scriptsize$\pm$4.8} & 0.975{\scriptsize$\pm$0.034} & 0.0255{\scriptsize$\pm$0.0273} & 0.952{\scriptsize$\pm$0.042} & 19.83{\scriptsize$\pm$5.0} & 0.986{\scriptsize$\pm$0.020} & 0.0158{\scriptsize$\pm$0.0167} & 0.955{\scriptsize$\pm$0.029} & 12.53 \\
\quad + $2\times$ steps & 21.68{\scriptsize$\pm$3.9} & 0.980{\scriptsize$\pm$0.021} & 0.0194{\scriptsize$\pm$0.0184} & 0.970{\scriptsize$\pm$0.022} & 22.77{\scriptsize$\pm$6.5} & 0.989{\scriptsize$\pm$0.012} & 0.0120{\scriptsize$\pm$0.0117} & 0.969{\scriptsize$\pm$0.022} & 8.81 \\
\midrule
FT + $\mathcal{L}_{\text{align}}$ & & & & & & & & & \\
\quad $\lambda = 10^{-3}$ & 22.49{\scriptsize$\pm$3.8} & 0.983{\scriptsize$\pm$0.017} & 0.0182{\scriptsize$\pm$0.0162} & 0.971{\scriptsize$\pm$0.021} & 22.89{\scriptsize$\pm$6.4} & 0.990{\scriptsize$\pm$0.011} & 0.0116{\scriptsize$\pm$0.0110} & 0.970{\scriptsize$\pm$0.021} & 8.79 \\
\quad $\lambda = 5\times 10^{-3}$ (*) & 22.63{\scriptsize$\pm$3.6} & 0.983{\scriptsize$\pm$0.018} & 0.0182{\scriptsize$\pm$0.0160} & 0.970{\scriptsize$\pm$0.023} & 22.48{\scriptsize$\pm$5.3} & 0.990{\scriptsize$\pm$0.011} & 0.0117{\scriptsize$\pm$0.0110} & 0.969{\scriptsize$\pm$0.021} & 8.94 \\
\quad $\lambda = 10^{-2}$ & 21.75{\scriptsize$\pm$4.2} & 0.981{\scriptsize$\pm$0.018} & 0.0193{\scriptsize$\pm$0.0157} & 0.969{\scriptsize$\pm$0.024} & 21.73{\scriptsize$\pm$6.0} & 0.988{\scriptsize$\pm$0.012} & 0.0127{\scriptsize$\pm$0.0112} & 0.966{\scriptsize$\pm$0.023} & 9.64 \\
\midrule
\rowcolor{gray!10} Ours (w.r.t. $\mathbf{x}_t$) (*) & 25.26{\scriptsize$\pm$3.9} & 0.985{\scriptsize$\pm$0.017} & 0.0157{\scriptsize$\pm$0.0152} & 0.979{\scriptsize$\pm$0.017} & 25.91{\scriptsize$\pm$7.2} & \underline{0.991{\scriptsize$\pm$0.011}} & 0.0105{\scriptsize$\pm$0.0109} & 0.974{\scriptsize$\pm$0.021} & 7.41 \\
\rowcolor{gray!10} Ours (w.r.t. $\hat{\mathbf{x}}_1$) (*) & 26.40{\scriptsize$\pm$3.8} & 0.987{\scriptsize$\pm$0.016} & 0.0140{\scriptsize$\pm$0.0142} & 0.983{\scriptsize$\pm$0.014} & \textbf{26.53{\scriptsize$\pm$7.2}} & \textbf{0.992{\scriptsize$\pm$0.010}} & \underline{0.0098{\scriptsize$\pm$0.0101}} & \underline{0.977{\scriptsize$\pm$0.019}} & \underline{6.64} \\
\rowcolor{gray!10} \quad + CFG ($w = 3$) & \underline{27.02{\scriptsize$\pm$5.1}} & \textbf{0.989{\scriptsize$\pm$0.013}} & \textbf{0.0124{\scriptsize$\pm$0.0128}} & \textbf{0.986{\scriptsize$\pm$0.017}} & \underline{25.85{\scriptsize$\pm$7.2}} & \textbf{0.992{\scriptsize$\pm$0.010}} & \textbf{0.0096{\scriptsize$\pm$0.0100}} & \textbf{0.978{\scriptsize$\pm$0.018}} & \textbf{6.25} \\
\rowcolor{gray!10} \quad + CFG ($w = 5$) & 26.35{\scriptsize$\pm$5.6} & \textbf{0.989{\scriptsize$\pm$0.012}} & \underline{0.0139{\scriptsize$\pm$0.0153}} & \underline{0.983{\scriptsize$\pm$0.023}} & 24.43{\scriptsize$\pm$7.2} & 0.991{\scriptsize$\pm$0.011} & 0.0109{\scriptsize$\pm$0.0111} & 0.973{\scriptsize$\pm$0.022} & 7.35 \\
\bottomrule
\end{tabular}%
}
\end{table*}

\begin{table*}[t]
\centering
\caption{\textbf{Extended Quantitative Results for 3D Texturing (Toys4K).} Asterisks (*) denote base configurations compared in the main text. Shaded rows indicate \algname variants.}
\label{tab:3d_results_toys4k_extended}
\adjustbox{max width=\linewidth}{%
\begin{tabular}{l cccc ccccc}
\toprule
& \multicolumn{4}{c}{\textbf{Fidelity}} & \multicolumn{5}{c}{\textbf{Plausibility}} \\
\cmidrule(lr){2-5} \cmidrule(lr){6-10}
Method & M. PSNR $\uparrow$ & SSIM $\uparrow$ & LPIPS $\downarrow$ & CLIP $\uparrow$ & MV-M.PSNR $\uparrow$ & MV-SSIM $\uparrow$ & MV-LPIPS $\downarrow$ & MV-CLIP $\uparrow$ & FID $\downarrow$ \\
\midrule
IT Guidance (*) &  26.82{\scriptsize$\pm$3.2} & 0.990{\scriptsize$\pm$0.009} & 0.0137{\scriptsize$\pm$0.0136} & 0.982{\scriptsize$\pm$0.017} & 23.63{\scriptsize$\pm$4.2} & \underline{0.992{\scriptsize$\pm$0.008}} & 0.0112{\scriptsize$\pm$0.0111} & 0.972{\scriptsize$\pm$0.020} & 8.43 \\
\quad + $2\times$ steps &  27.95{\scriptsize$\pm$3.2} & 0.991{\scriptsize$\pm$0.008} & 0.0127{\scriptsize$\pm$0.0125} & 0.984{\scriptsize$\pm$0.016} & 24.42{\scriptsize$\pm$4.7} & \underline{0.992{\scriptsize$\pm$0.009}} & 0.0111{\scriptsize$\pm$0.0110} & 0.973{\scriptsize$\pm$0.020} & 7.90 \\
\quad + $s = 0.5$ &  \underline{28.02{\scriptsize$\pm$3.4}} & 0.991{\scriptsize$\pm$0.009} & 0.0163{\scriptsize$\pm$0.0195} & 0.975{\scriptsize$\pm$0.030} & 23.94{\scriptsize$\pm$5.2} & 0.991{\scriptsize$\pm$0.008} & 0.0120{\scriptsize$\pm$0.0116} & 0.970{\scriptsize$\pm$0.020} & 9.00 \\
\midrule
Standard FT (*) & 23.00{\scriptsize$\pm$4.2} & 0.986{\scriptsize$\pm$0.014} & 0.0160{\scriptsize$\pm$0.0164} & 0.975{\scriptsize$\pm$0.024} & 24.38{\scriptsize$\pm$6.9} & 0.991{\scriptsize$\pm$0.010} & 0.0114{\scriptsize$\pm$0.0118} & 0.972{\scriptsize$\pm$0.021} & 8.87 \\
\quad + CFG ($w = 3$) & 21.43{\scriptsize$\pm$4.4} & 0.985{\scriptsize$\pm$0.015} & 0.0182{\scriptsize$\pm$0.0177} & 0.968{\scriptsize$\pm$0.023} & 21.26{\scriptsize$\pm$4.9} & 0.990{\scriptsize$\pm$0.010} & 0.0130{\scriptsize$\pm$0.0124} & 0.965{\scriptsize$\pm$0.023} & 10.30 \\
\quad + CFG ($w = 5$) & 20.55{\scriptsize$\pm$4.6} & 0.984{\scriptsize$\pm$0.016} & 0.0210{\scriptsize$\pm$0.0214} & 0.963{\scriptsize$\pm$0.025} & 20.32{\scriptsize$\pm$4.9} & 0.989{\scriptsize$\pm$0.011} & 0.0151{\scriptsize$\pm$0.0149} & 0.960{\scriptsize$\pm$0.027} & 12.41 \\
\quad + $2\times$ steps & 22.66{\scriptsize$\pm$4.3} & 0.986{\scriptsize$\pm$0.015} & 0.0163{\scriptsize$\pm$0.0161} & 0.975{\scriptsize$\pm$0.024} & 23.04{\scriptsize$\pm$6.3} & 0.990{\scriptsize$\pm$0.010} & 0.0118{\scriptsize$\pm$0.0118} & 0.971{\scriptsize$\pm$0.020} & 8.75 \\
\midrule
FT + $\mathcal{L}_{\text{align}}$ & & & & & & & & & \\
\quad $\lambda = 10^{-3}$ & 23.01{\scriptsize$\pm$4.1} & 0.986{\scriptsize$\pm$0.014} & 0.0162{\scriptsize$\pm$0.0173} & 0.975{\scriptsize$\pm$0.023} & 23.39{\scriptsize$\pm$5.3} & 0.991{\scriptsize$\pm$0.010} & 0.0114{\scriptsize$\pm$0.0124} & 0.972{\scriptsize$\pm$0.022} & 7.99 \\
\quad $\lambda = 5\times 10^{-3}$ (*) & 23.05{\scriptsize$\pm$4.0} & 0.986{\scriptsize$\pm$0.014} & 0.0159{\scriptsize$\pm$0.0162} & 0.976{\scriptsize$\pm$0.023} & 23.24{\scriptsize$\pm$5.0} & 0.991{\scriptsize$\pm$0.009} & 0.0111{\scriptsize$\pm$0.0112} & 0.973{\scriptsize$\pm$0.022} & 8.30 \\
\quad $\lambda = 10^{-2}$ & 22.79{\scriptsize$\pm$4.2} & 0.986{\scriptsize$\pm$0.015} & 0.0168{\scriptsize$\pm$0.0169} & 0.975{\scriptsize$\pm$0.024} & 23.16{\scriptsize$\pm$5.7} & 0.990{\scriptsize$\pm$0.010} & 0.0119{\scriptsize$\pm$0.0116} & 0.971{\scriptsize$\pm$0.022} & 8.60 \\
\midrule
\rowcolor{gray!10} Ours (w.r.t. $\mathbf{x}_t$) (*) & 26.54{\scriptsize$\pm$4.2} & 0.990{\scriptsize$\pm$0.011} & 0.0129{\scriptsize$\pm$0.0140} & 0.980{\scriptsize$\pm$0.024} & \underline{27.20{\scriptsize$\pm$8.1}} & \textbf{0.993{\scriptsize$\pm$0.008}} & 0.0097{\scriptsize$\pm$0.0104} & 0.977{\scriptsize$\pm$0.020} & 7.20 \\
\rowcolor{gray!10} Ours (w.r.t. $\hat{\mathbf{x}}_1$) (*) & 27.26{\scriptsize$\pm$4.4} & 0.991{\scriptsize$\pm$0.010} & 0.0116{\scriptsize$\pm$0.0125} & 0.984{\scriptsize$\pm$0.022} & \textbf{27.21{\scriptsize$\pm$7.5}} & \textbf{0.993{\scriptsize$\pm$0.008}} & \textbf{0.0092{\scriptsize$\pm$0.0100}} & \underline{0.978{\scriptsize$\pm$0.021}} & \underline{6.66} \\
\rowcolor{gray!10} \quad + CFG ($w = 3$) & \textbf{28.30{\scriptsize$\pm$4.0}} & \textbf{0.993{\scriptsize$\pm$0.006}} & \textbf{0.0099{\scriptsize$\pm$0.0102}} & \textbf{0.990{\scriptsize$\pm$0.010}} & 26.18{\scriptsize$\pm$6.7} & \textbf{0.993{\scriptsize$\pm$0.008}} & \underline{0.0093{\scriptsize$\pm$0.0102}} & \textbf{0.979{\scriptsize$\pm$0.018}} & \textbf{6.01} \\
\rowcolor{gray!10} \quad + CFG ($w = 5$) & 27.44{\scriptsize$\pm$4.8} & \underline{0.992{\scriptsize$\pm$0.007}} & \underline{0.0114{\scriptsize$\pm$0.0120}} & \underline{0.986{\scriptsize$\pm$0.017}} & 24.26{\scriptsize$\pm$6.7} & \underline{0.992{\scriptsize$\pm$0.008}} & 0.0109{\scriptsize$\pm$0.0120} & 0.974{\scriptsize$\pm$0.022} & 7.53 \\
\bottomrule
\end{tabular}%
}
\end{table*}

\clearpage

\end{document}